\documentclass[letterpaper,11pt]{article}
\usepackage{paper}

\usepackage{amsmath}    
\usepackage{graphicx}   
\usepackage{booktabs}   
\usepackage{longtable}  
\usepackage{url}        
\usepackage{float}      
\usepackage{array}      
\usepackage{tabularx}   
\newcolumntype{Y}{>{\raggedright\arraybackslash}X}

\newcommand{\filepath}[1]{{\small\ttfamily #1}}

\hypersetup{pdftitle={A Neural Affinity Framework for Abstract Reasoning: Diagnosing the Compositional Gap in Transformer Architectures via Procedural Task Taxonomy}}

\begin{document}

\title{A Neural Affinity Framework for Abstract Reasoning: Diagnosing the Compositional Gap in Transformer Architectures via Procedural Task Taxonomy}

\author{Miguel Ingram\thanks{M. Ingram: Corresponding author. Email: miguel.ingram.research@gmail.com. Independent researcher. Conducted all experimental analysis and wrote the manuscript.} 
\and 
Arthur Merritt\thanks{A. Merritt: Independent researcher. Designed the champion model architecture.}}

\date{November 2025}   

\available{https://github.com/ImmortalDemonGod/arc-taxonomy-reproduction}

\maketitle
\vspace{-1.2em}


\vspace{0.5em}

\begin{abstract}
\setlength{\parskip}{0.2em}
\setlength{\baselineskip}{0.92\baselineskip}
\footnotesize
Responding to Hodel et al.'s (2024) call for a formal definition of task relatedness in \texttt{re-arc}, we present the first \textbf{9-category taxonomy} of all 400 tasks, validated at 97.5\% accuracy via rule-based code analysis. We prove the taxonomy's visual coherence by training a CNN on raw grid pixels (\textbf{95.24\% accuracy on S3, 36.25\% overall, 3.3$\times$ chance}), then apply the taxonomy diagnostically to the original ARC-AGI-2 test set. Our curriculum analysis reveals \textbf{35.3\%} of tasks exhibit low \textbf{neural affinity} for Transformers—a distributional bias mirroring ARC-AGI-2. To probe this misalignment, we fine-tuned a 1.7M-parameter Transformer across 302 tasks, revealing a profound \textbf{Compositional Gap}: \textbf{210 of 302 tasks (69.5\%) achieve >80\% cell accuracy (local patterns) but <10\% grid accuracy (global synthesis)}. This provides direct evidence for a \textbf{Neural Affinity Ceiling Effect}, where performance is bounded by architectural suitability, not curriculum. \textbf{Applying our framework to Li et al.'s independent ViTARC study (400 specialists, 1M examples each) confirms its predictive power: Very Low affinity tasks achieve 51.9\% versus 77.7\% for High affinity (p<0.001), with a task at 0\% despite massive data.} The taxonomy enables precise diagnosis: low-affinity tasks (A2) hit hard ceilings, while high-affinity tasks (C1) reach 99.8\%. These findings indicate that progress requires \textbf{hybrid architectures} with affinity-aligned modules. \textbf{We release our validated taxonomy, classifiers, and reproduction package.}
\end{abstract}
\vspace{-0.8em}

\section{Introduction}

The Abstraction and Reasoning Corpus (ARC) \citep{Chollet2019b} presents a formidable and enduring challenge to modern artificial intelligence. It was designed not merely to test performance but to measure a more general, human-like form of fluid intelligence that transcends mere pattern recognition—a goal that remains largely unmet. The persistent chasm between human intuition and machine performance on ARC highlights a fundamental misalignment between the dominant Transformer architecture and the demands of abstract reasoning. This suggests that progress is not merely a matter of scaling but requires a deeper, more principled approach to architectural design. This paper introduces a systematic, empirical framework for diagnosing this misalignment and charting a path toward more capable reasoning systems.

While state-of-the-art methods have made progress, they often function as complex engineering solutions, leaving a critical \textbf{diagnostic gap}: the community lacks a systematic methodology to understand \textit{why} models fail on some tasks and succeed on others. Without a formal vocabulary for task types and their corresponding architectural requirements, research risks devolving into a cycle of brute-force scaling and ad-hoc engineering. Our work fills this gap by providing a comprehensive diagnostic toolkit, enabling researchers to move from simply measuring performance to understanding the root causes of architectural failure and success.

To fill this diagnostic gap in a principled way, our methodology directly implements the theoretical framework for measuring intelligence articulated in Chollet's ``On the Measure of Intelligence'' \citep{Chollet2019b}. This framework posits that true intelligence is not skill at known tasks but rather the \textit{skill-acquisition efficiency} of a system when faced with novelty, an efficiency that is deeply dependent on its \textit{priors} (innate knowledge) and the \textit{generalization difficulty} of the tasks. Our approach operationalizes each of these concepts to create a fair and insightful measure of architectural capability. Specifically, we use large-scale pre-training on \texttt{re-arc} not to "buy skill" on the test set, but to \textbf{forge the Core Knowledge priors} (e.g., objectness, topology, geometry) that ARC assumes but which a "blank slate" Transformer lacks. With this prior-endowed model as a fair baseline, our fine-tuning experiments then directly measure \textbf{skill-acquisition efficiency} on novel tasks. Finally, our taxonomy serves to \textbf{deconstruct generalization difficulty} into a spectrum of measurable computational primitives. This alignment transforms what might appear as "teaching the test" into a rigorous, philosophically-grounded measurement of architectural affinity.

Built on this foundation, our primary contribution is a two-part diagnostic framework. First, we developed the first systematic \textbf{9-category taxonomy} of all 400 \texttt{re-arc} tasks \citep{Hodel2024}, providing a formal language for "task relatedness." A critical and valid concern with any taxonomy derived from programmatic generator code is that it may not generalize to the visual domain of ARC itself. We address this head-on with a rigorous two-stage validation. The taxonomy is first validated with \textbf{97.5\% accuracy} by a rule-based classifier on the generator code. We then prove its relevance to the visual domain by training a separate CNN-based classifier on raw \texttt{re-arc} grid pairs. This purely visual classifier successfully learns to distinguish our categories from pixel data alone—achieving \textbf{95.24\% accuracy on the dominant S3 (Topological) category}—and can subsequently apply these labels to tasks from the \textbf{original ARC-AGI-2 benchmark}, confirming our framework's applicability to human-designed problems.

With this validated "map" of the ARC problem space, we demonstrate its diagnostic power. Our first application is to the \texttt{re-arc} curriculum itself, revealing a significant structural bias: \textbf{35.3\%} of tasks fall into categories with low documented neural affinity for standard Transformers. Our second, more crucial application is to diagnose the performance of our 1.7M parameter champion model. This analysis uncovers the central empirical finding of this paper: a profound \textbf{"Compositional Gap."} We define this as the stark dissociation between a model's ability to learn local patterns (measured by cell-level accuracy) and its inability to compose them into a globally correct solution (grid-level accuracy). The evidence for this gap is overwhelming: after extensive, task-specific fine-tuning on \textbf{302 tasks}, \textbf{210 tasks (69.5\%)} exhibit high local understanding (>80\% cell accuracy) while simultaneously failing to achieve global correctness (<10\% grid accuracy). This pattern, where the model masters the parts but fails the whole, provides a mechanistic explanation for the performance ceilings observed in ARC.

Ultimately, the value of any framework developed on \texttt{re-arc} must be measured by its ability to explain performance on the original, human-designed ARC benchmark. Therefore, our analysis culminates by applying our best pre-trained model to the \textbf{ARC-AGI-2 test set}. We confront the synthetic-to-real \textbf{generalization gap} directly, where grid accuracy drops precipitously from \textbf{2.34\% on \texttt{re-arc} to 0.34\% (95\% CI: [0.18\%, 0.49\%]) on ARC-AGI-2}. Crucially, we then deploy our taxonomy as a diagnostic tool to explain this failure. Our analysis of the ARC-AGI-2 test set reveals that the most visually structured tasks—those for which our classifier is most confident—are \textbf{100\% concentrated in the S2 and S3 categories}. Given that these categories are known from our \texttt{re-arc} analysis to be challenging for compositional synthesis, this strongly suggests that a significant portion of ARC-AGI-2's core difficulty lies in these low-to-medium affinity domains, validating our taxonomy's predictive power on real-world generalization failures.

To test whether our Neural Affinity Framework generalizes beyond our own model, we \textbf{apply it as a predictive lens to ViTARC} \citep{Li2024}, an independent study that trained 400 specialist Vision Transformers (one per task) with up to 1 million examples each. Despite architectural enhancements (2D-aware representations, object-based positional encodings) and massive data, our framework successfully predicts the observed performance patterns: \textbf{Very Low affinity tasks (A1, A2) achieve only 51.9\% mean solve rate versus 77.7\% for High affinity tasks (C1)}—a statistically significant 25.8 percentage point gap (p < 0.001, Cohen's d = 0.726). Critically, their results include a "smoking gun" task (137eaa0f, A2: Spatial Packing) that achieves \textbf{0.0\% success despite 1 million training examples}, providing definitive evidence that the ceiling is architectural, not data-limited.

\begin{samepage}
\noindent\rule{\linewidth}{0.4pt}
\begin{center}
  \textbf{\large Core Thesis}
\end{center}
\begin{quote}\itshape
The persistent failure of standard Transformers on ARC is not a matter of scale, data, or curriculum interference, but a fundamental architectural misalignment. By systematically classifying the ARC curriculum—and validating that our classification applies to the visual tasks themselves—we empirically prove that a \textbf{``Compositional Gap''} is the primary failure mode. This gap, which explains the \textbf{``Neural Affinity Ceiling Effect,''} persists from the synthetic \texttt{re-arc} domain to the real ARC-AGI-2 benchmark, where our taxonomy serves as a powerful diagnostic tool for predicting and explaining generalization failures.
\end{quote}
\noindent\rule{\linewidth}{0.4pt}
\end{samepage}

This diagnostic framework also contextualizes recent advances in the field. For instance, reported gains from architectures incorporating recurrence (e.g., TRM~\cite{JolicoeurMartineau2025}) can be interpreted through our lens as affinity-aligned modules targeted at iterative (A1) deficits. Our framework thus offers a principled hypothesis for why such designs help—recurrence supplies the learned iteration that fixed-depth Transformers lack—while we defer a full, systematic validation to future work.

\textbf{This paper makes the following contributions:}

\begin{enumerate}
  \item \textbf{A Systematic and Dually-Validated Taxonomy:} We introduce the first 9-category taxonomy for \texttt{re-arc}, validated at both the code-level (97.5\% accuracy) and the visual-level (95.24\% on S3, 36.25\% overall 9-way), and demonstrate its applicability to the original ARC-AGI-2 benchmark.
  \item \textbf{Quantitative Evidence for Curriculum Bias:} We provide the first systematic analysis of the \texttt{re-arc} curriculum, revealing that 35.3\% of tasks fall into low neural affinity categories for standard Transformers.
  \item \textbf{Empirical Evidence for the Compositional Gap:} We provide the first quantitative evidence for the "Compositional Gap" as the primary failure mode for standard Transformers on ARC, where 210 of our 302 fine-tuned tasks (69.5\%) achieve >80\% cell-level accuracy but <10\% grid-level accuracy.
  \item \textbf{A Diagnostic Framework for Architectural Ceilings:} We introduce the Neural Affinity Ceiling Effect as a mechanism to explain these failures and demonstrate its predictive power on the ARC-AGI-2 generalization gap.
  \item \textbf{Predictive Power on Independent Data:} We demonstrate our framework successfully predicts performance patterns in Li et al.'s independent ViTARC study (400 specialist models, 1M examples each), where Very Low affinity tasks achieve 25.8 percentage points lower performance than High affinity tasks (p < 0.001), including a task at 0\% despite massive data.
  \item \textbf{Evidence for Transfer Despite the Gap:} We demonstrate that pre-training on \texttt{re-arc} improves ARC-AGI-2 surrogate metrics (cell accuracy: ~71.6\% $\rightarrow$ ~89.5\%), proving that priors transfer even when compositional synthesis fails.
  \item \textbf{A Publicly Released Toolkit:} We release our validated taxonomy, classifiers, datasets, and reproduction package to accelerate community research on architectural diagnostics.
\end{enumerate}

\textbf{A Critical Methodological Note on Scope:} Our taxonomy is fundamentally a classification of the groups of computational primitives implemented in the \texttt{re-arc} procedural generators—a systematic reverse-engineering of Hodel et al.'s design choices. While our dual validation (code + visual) and external transfer experiments demonstrate that these categories capture meaningful patterns that generalize to human-designed ARC tasks, the taxonomy is contingent on the specific implementation of \texttt{re-arc}. A different procedural generation effort might yield different categorical boundaries. We therefore position this work as providing a validated taxonomy of the \texttt{re-arc} distribution, with strong empirical evidence (via visual classifier transfer to ARC-AGI-2 and external validation on ViTARC) that it captures broadly relevant reasoning primitives, rather than claiming it is a definitive, universal decomposition of abstract reasoning itself.

The remainder of this paper is structured as follows: Section 2 establishes the theoretical foundations of neural affinity and architectural limitations; Section 3 provides background on ARC and \texttt{re-arc}; Section 4 reviews related work; Section 5 presents our taxonomy development and validation; Section 6 describes our experimental methodology; Section 7 reports our empirical findings on the compositional gap and neural affinity ceiling effect; Section 8 discusses implications and limitations; and Section 9 concludes.

\section{A Primer on Neural Affinity and the Transformer Architecture}

This section synthesizes the architectural foundations that motivate our taxonomy. Comprehensive literature citations follow in Section 4.

\subsection{The Transformer's Core Trade-Off: Global Reach vs. Local Blindness}

The Transformer’s central innovation, self-attention \citep{Vaswani2017}, provides a global receptive field from its first layer by establishing a fully connected graph where the path length between any two tokens is constant. This design grants the architecture an intrinsic affinity for tasks defined by long-range context. This capability manifests as specific, interpretable mechanisms, such as the global "association" attention pattern, where heads learn to link all occurrences of an identical token based on content, irrespective of position \citep{Zhang2022}. 

 This power, however, is predicated on a foundational trade-off: in exchange for powerful global aggregation, the architecture discards the strong local inductive biases inherent to models like CNNs. The resulting design tension is starkly evident in grid-based domains like ARC, where vanilla Vision Transformers (ViTs) fail catastrophically until explicit 2D-aware structural priors are re-introduced to ground the global attention mechanism \citep{Li2024}. Even on synthetic reasoning tasks, successful models must evolve distinct attention heads for global ``association'' versus local ``manipulation,'' demonstrating that local processing is a learned compensation, not a native capability \citep{Zhang2022}. This highlights a fundamental dilemma for grid-based reasoning tasks like ARC: the Transformer's power in global aggregation is predicated on its architectural indifference to the local, structured feature synthesis required for spatial problem-solving.
As further evidenced by analyses that interpret attention heads as implicit graph processors, structural priors can remediate this tension by providing relational scaffolds when local structure is essential \citep{Sanford2024}.

\subsection{Algorithmic Deficiencies: The Mismatch with Symbolic \& Sequential Reasoning}

This architectural trade-off leads to fundamental algorithmic deficiencies when tasks require symbolic precision or sequential state management—both hallmarks of ARC. A Transformer with a fixed depth of \texttt{L} layers is limited to approximately \texttt{L} parallelizable computational steps, rendering it architecturally unsuited for algorithms requiring a dynamic or unbounded number of iterations \citep{Chen2024}. This is not a theoretical curiosity but an empirical reality. Mechanistic analysis of Transformers trained on graph search tasks reveals they do not learn iterative algorithms like Depth-First Search. Instead, they converge on a brittle, fixed-depth parallel "path-merging" algorithm that fails to generalize as problem size and required search depth grow \citep{Saparov2024}. The model does not learn to iterate; it learns a shallow simulacrum of the algorithm that breaks on longer problem instances.

This inability to execute robust, multi-step procedures is compounded by architectural features that disrupt symbolic precision—especially for counting. In particular, \texttt{softmax} normalizes attention weights to sum to one, erasing absolute magnitude (e.g., two vs. five items), and \texttt{LayerNorm} rescales activations, further disturbing precise numerical values \citep{Ouellette2023}. See also \citep{Behrens2025}, which isolates how attention and feed-forward layers interplay to enable or hinder counting. The RASP-Generalization Conjecture offers a unifying framework for these failures, positing that Transformers are unlikely to learn inherently serial algorithms—from graph search to parity checking—that lack short, parallelizable programs in the Transformer's native computational model of selection and aggregation \citep{Zhou2023}. Relatedly, attempts to learn iterative cellular dynamics further highlight fixed-depth limitations \citep{Burtsev2024}. This demonstrates that the deficiency is not one of learning capacity, but a fundamental mismatch between classical serial algorithms and the Transformer's native computational primitives.

\subsection{Synthesizing the Transformer's Theoretical Neural Affinity for ARC}

This body of evidence reveals a clear hierarchy of neural affinity for the computational primitives found in ARC, which we formalize into four levels:

In this hierarchy, \textbf{High Affinity} corresponds to tasks solvable via global pattern matching and information aggregation, leveraging the Transformer's native associative retrieval capabilities; \textbf{Medium Affinity} captures tasks that require multi-step composition of these primitives through the model's layered structure, introducing moderate difficulty; \textbf{Low Affinity} comprises tasks demanding precise local manipulation, graph-like reasoning, or symbolic precision—domains hindered by architectural components like \texttt{softmax} and \texttt{LayerNorm}; and \textbf{Very Low Affinity} denotes tasks requiring iterative algorithms or combinatorial search, which are fundamentally limited by the model's fixed-depth, non-recurrent architecture.

This theoretical hierarchy, summarized in the table below, motivates our 9-category taxonomy and the Neural Affinity Framework. It allows us to map specific ARC task categories to their expected difficulty for a standard Transformer, providing a predictive lens through which to analyze our empirical results.

\begin{table}[t]
  \centering
  \caption{Theoretical neural affinity levels and associated task primitives}
  \label{tab:affinity-levels}
  \begin{tabular}{@{}ll@{}}
    \toprule
    \textbf{Affinity Level} & \textbf{Associated Task Primitives} \\
    \midrule
    High & Local color/pattern ops; global spatial transforms \\
    Medium & Multi-step composition of known primitives \\
    Low & Graph-like/topological reasoning (S3); iterative (A1) \\
    Very Low & Combinatorial search (A2) \\
    \bottomrule
  \end{tabular}
\end{table}

With this theoretical framework for architectural affinity established, we now turn to the specific problem domain where we will test its predictive power: the Abstraction and Reasoning Corpus.

\section{Background}

\subsection{The Abstraction and Reasoning Corpus (ARC): A Benchmark for Fluid Intelligence}

The Abstraction and Reasoning Corpus (ARC) \citep{Chollet2019b} was introduced not merely as another benchmark, but as a direct challenge to the prevailing methods of evaluating artificial intelligence. Its design is deeply rooted in a philosophical framework articulated in "On the Measure of Intelligence" \citep{Chollet2019b}, which posits that true intelligence cannot be measured by skill on known tasks, but must instead be gauged by the \textbf{efficiency of skill acquisition in the face of novelty}.

Chollet argues that much of the perceived progress in AI has resulted from "buying skill" with massive datasets and computational power, a process reliant on interpolation within a known data distribution rather than genuine, flexible reasoning. To counteract this, ARC was designed to be trivial for humans—who possess strong, generalizable priors—but exceptionally difficult for AI systems that lack these foundational abilities. The benchmark's core mechanics are engineered to enforce this distinction:

\begin{itemize}
  \item \textbf{Few-shot learning.} Each task provides only a handful of demonstration pairs (typically 2--5). This data-sparse setting renders methods based on statistical pattern matching ineffective, forcing abstraction and rule induction.
  \item \textbf{Novelty.} The evaluation set contains tasks that are unique and unknown in advance, a critical control that measures a system's ability to handle ``unknown unknowns'' rather than performance on a familiar distribution \citep[p.~45]{Chollet2019b}.
  \item \textbf{Core Knowledge priors.} Tasks are solvable using intuitive priors humans are believed to possess innately, such as \textit{objectness} (parsing a scene into discrete objects), \textit{goal-directedness}, \textit{numbers and counting}, and \textit{basic geometry and topology} (e.g., symmetry, connectivity) \citep[p.~47--50]{Chollet2019b}. Failures on ARC often indicate missing fundamental world-models.
  \item \textbf{Algorithmic precision.} The pixel-perfect output requirement transforms the task from plausible generation into precise, algorithmic execution. A solver cannot merely produce an output that ``looks right''; it must implement the inferred transformation exactly.
\end{itemize}

This design makes ARC a powerful instrument for measuring a system's \textbf{fluid intelligence}—its ability to reason and solve novel problems independent of acquired knowledge. The persistent gap between human performance and that of even the largest AI models highlights a fundamental architectural misalignment. Our work directly engages with this challenge by operationalizing Chollet's framework: we use \texttt{re-arc} to forge the necessary \textbf{priors}, we use our taxonomy to deconstruct the \textbf{generalization difficulty} of ARC tasks, and we use fine-tuning experiments to measure \textbf{skill-acquisition efficiency}. This methodology allows us to systematically diagnose the sources of this architectural misalignment.

\subsection{The \texttt{re-arc} Dataset and Hodel's Calls to Action}

The few-shot constraint of the original ARC-AGI-1 benchmark, while effective for testing human-like generalization, made large-scale controlled experiments infeasible. Hodel et al. (2024) addressed this limitation by introducing \texttt{re-arc}, a procedurally generated dataset containing unlimited examples for all 400 ARC training tasks. By reverse-engineering the transformation logic—what Hodel terms the "spirit" of each task—into a programmatic generator, \texttt{re-arc} transforms ARC from a static benchmark into a dynamic experimental testbed, enabling "fundamental scientific experiments" that were previously impossible.

Hodel's paper articulates a research program through four specific calls to action, each of which this work systematically addresses:

1. \textbf{Conduct within-task generalization experiments.} Hodel proposes "comparing various model architectures...in a setting of having a separate model trained for each task: For a given fixed number of examples, how well can the model generalize to an unseen set of examples of the same task?" [Hodel et al., 2024, Section 1]. We directly implement this experiment at scale: our task-specific fine-tuning protocol (Section 6.4) trains 302 individual models via LoRA on 400 examples each and evaluates on held-out examples, measuring within-task generalization across the full task distribution.

2. \textbf{Gather important insights into how progress could be made.} While acknowledging that within-task experiments might not yield "strong statements about suitability of approaches," Hodel argues they "may [be] a means to gather important insights" [Hodel et al., 2024, Section 1]. We embrace this diagnostic philosophy: rather than pursuing state-of-the-art scores, we identify the Compositional Gap as the primary failure mode and the Neural Affinity Ceiling Effect as its mechanism, providing actionable insights for architectural innovation (Sections 7-8).

3. \textbf{Enable non-uniform curricula construction.} Hodel explicitly calls for improving sample efficiency through "constructing a curriculum that is not fully random or uniform in terms of some metric" [Hodel et al., 2024, Section 1], even proposing preliminary metrics like \texttt{RNG-Difficulty} and \texttt{PSO-Difficulty} as "inspiration" [Hodel et al., 2024, Section 5]. Our 9-category taxonomy and Neural Affinity Framework provide a principled, empirically validated metric system for this purpose, with dual validation at both the code level (97.5\% accuracy) and the visual level (36.25\% overall 9-way accuracy; 95.24\% on S3). Our curriculum analysis (Section 5.5) then reveals that 35.3\% of \texttt{re-arc} falls into low-affinity categories, providing the empirical foundation for stratified curriculum design (Section 8.2).

4. \textbf{Define task relatedness for across-task generalization.} Hodel distinguishes within-task generalization from "the much more relevant notion of across-task generalization: Can a model also solve different (but related) tasks?" [Hodel et al., 2024, Section 5], implicitly calling for a formal definition of "relatedness." Our taxonomy is the first empirically validated answer to this call, defining relatedness through shared computational primitives. Following Hodel's principle of using verifiers to establish credibility [Hodel et al., 2024, Section 4], we apply dual validation—first at the code level, then at the visual level—demonstrating generalizability by successfully labeling the human-designed ARC-AGI-2 benchmark (Section 5.3).

Section 8 provides a detailed discussion of how each contribution fulfills these calls and advances Hodel's vision for systematic, scientific progress on ARC.

\section{Related Work}

\subsection{Transformer Architectural Limitations}

A model's performance ceiling on any given task is fundamentally constrained by its architectural suitability for the computational primitives required by that task. Our Neural Affinity Framework provides a lens to analyze this constraint. This section justifies our affinity ratings by synthesizing evidence from the literature to argue for three critical weaknesses in the standard Transformer architecture, which directly correspond to the low-affinity categories in our taxonomy: (1) a misalignment with structured spatial and graph-based reasoning (S3), (2) an inability to learn robust, iterative algorithms (A1/A2), and (3) architectural barriers to symbolic precision and counting (L1).

\subsubsection{Weaknesses in Spatial and Graph-based Reasoning (Justifying S3's Low Affinity)}

The Transformer's core design trades local inductive biases for a global receptive field, making it inherently ill-suited for tasks demanding precise local and topological reasoning. The architecture's central innovation, self-attention, grants it a global receptive field from its first layer by establishing a fully connected graph over its input tokens (Vaswani et al., 2017). However, this comes at a steep price: the forfeiture of the strong, built-in local processing capabilities of architectures like CNNs. As a deep mechanistic analysis by Raghu et al. (2021) reveals, Vision Transformers (ViTs) must learn local information processing from scratch, a capability that is hard-coded into CNNs and only emerges with massive-scale pre-training data. This makes local reasoning a data-hungry, non-native capability for the Transformer.

This architectural "blindness" to local structure leads to catastrophic empirical failures on abstract reasoning tasks like ARC. Li et al. (2025) provide direct evidence of this, showing that a vanilla ViT \textbf{"fails dramatically on most ARC tasks even when trained on one million examples per task,"} achieving a mean solve rate of only \textbf{17.68\%}. They attribute this failure to an "inherent representational deficiency" that prevents the model from capturing the "spatial relationships between the objects" and handling "complex visual structures"—the very essence of topological reasoning. This confirms that the Transformer's global aggregation mechanism is misaligned with the demands of local, grid-based problem-solving, providing a firm justification for the Low Affinity rating of our S3 (Topological) category.

This failure at the pixel-feature level suggests a deeper representational challenge, which the success of hybrid neuro-symbolic systems confirms. For instance, the ARGA framework (Xu et al., 2022) succeeds by first converting ARC grids into an explicit "graph of objects with spatial or other relations" and then performing a combinatorial search. By reframing the problem in a more natural, symbolic space, ARGA solves complex object-centric tasks with \textbf{~300 times fewer search nodes} than a leading Kaggle solution, suggesting that a graph-based abstraction is a more effective primitive for this reasoning domain than the Transformer's fully-connected attention graph over pixels.

This general difficulty with graph-like structures, however, is not uniform. As we will show in Section 7.4, it is most pronounced in tasks requiring true graph traversal and relational reasoning (our S3-B subcategory), while tasks with simpler, pattern-based topology (S3-A) prove more tractable for the Transformer architecture.

\subsubsection{Inability to Learn Iterative and Search-based Algorithms (Justifying A1/A2's Very Low Affinity)}

The Transformer's fixed-depth, non-recurrent architecture imposes a hard ceiling on its ability to perform tasks requiring a dynamic or unbounded number of sequential steps, such as iterative refinement or combinatorial search. Theoretically, a Transformer with L layers is limited to performing approximately L parallelizable computational steps in a single forward pass (Chen \& Zou, 2024). This makes it architecturally unsuited for algorithms that are inherently serial or require an unknown number of iterations to converge on a solution, as is common in our A1 (Iterative Dynamics) and A2 (Spatial Packing) categories, justifying their "Very Low" affinity rating.

This theoretical limit is confirmed with stark empirical and mechanistic evidence. When trained on a canonical graph search task, the Transformer does not learn an iterative algorithm like Depth-First Search. Instead, Saparov et al. (2025) find that it converges on a brittle, fixed-depth parallel "path-merging" algorithm that fails to generalize as the required search depth increases. Crucially, they found that increasing model parameters from \textbf{0.9M to 60.4M did not resolve this scaling failure}, proving it is an architectural bottleneck, not a matter of capacity. This provides a direct mechanistic explanation for the hard performance ceilings observed in tasks requiring combinatorial search: the model fundamentally fails to learn an iterative process.

This fundamental limitation is so widely recognized that successful applications of Transformers to iterative problems consistently rely on external scaffolding to impose a sequential loop. To solve Constraint Satisfaction Problems (CSPs) like Sudoku, which typically require \textbf{20 to 60 steps of reasoning} (Yang et al., 2023), the standard Transformer is insufficient. Success is only achieved by either making the architecture explicitly \textit{recurrent} (Yang et al., 2023) or by deploying a single-step "solution refiner" \textit{iteratively} at test time (Xu et al., 2025). This workaround pattern is definitive proof that the base architecture lacks the necessary iterative capability. The state-of-the-art on ARC itself provides the ultimate validation for this architectural deficit: the 2024 ARC Prize report reveals that all top-scoring solutions were compelled to abandon the single-pass, static inference paradigm, instead relying on methods that externalize iterative reasoning. Discrete program search performs explicit algorithmic steps, while Test-Time Training iterates via gradient-based refinement, proving that success on ARC requires a mechanism for iteration that the base Transformer architecture simply does not possess (Chollet et al., 2024).

\subsubsection{Architectural Barriers to Symbolic Precision and Counting (Justifying L1's Low Affinity)}

Finally, specific components of the Transformer architecture are fundamentally at odds with preserving and manipulating precise quantitative information, as required for symbolic tasks like counting. These are not emergent weaknesses but direct consequences of core design choices. Ouellette et al. (2023) provide a definitive mechanistic analysis, identifying \texttt{softmax} and \texttt{LayerNorm} as the primary culprits. The \texttt{softmax} function, applied to attention scores, normalizes weights to sum to one, thereby erasing absolute quantitative information and making it impossible to distinguish attending to two items versus five. \texttt{LayerNorm} further \textbf{"disturbs the propagation of precise numerical values by rescaling activations"} based on the statistical properties of each activation vector, again destroying absolute magnitude.

This architectural incompatibility means that counting is not a native capability. When Transformers do learn to perform a counting-like task, they must resort to complex, non-obvious workarounds. Behrens et al. (2025) show that even a simple histogram task requires a \textbf{"delicate interplay between attention and feed-forward layers,"} where the model learns one of two specific strategies: a "relation-based" approach using attention for pairwise comparisons or a more parameter-heavy "inventory-based" approach using the FFN as a key-value memory. The fact that a seemingly simple task requires such complex, learned machinery—which is highly sensitive to architectural hyperparameters—reinforces the conclusion that counting has a low native affinity for the Transformer architecture. This fundamental barrier to precise cardinality representation naturally extends to set-theoretic operations like intersection, union, and difference, which rely on the accurate accounting of elements, thus justifying the "Low" affinity rating for our L1 (Logic/Set) category.

\begin{table}[H]
  \centering
  \scriptsize
  \caption{Neural Affinity Ratings and Their Architectural Foundations}
  \label{tab:taxonomy-curriculum}
  \begin{tabularx}{\textwidth}{@{}Y Y Y Y Y@{}}
    \toprule
    \textbf{Category} & \textbf{Affinity} & \textbf{Architectural Limitation} & \textbf{Mechanism \& Key Failure Mode} & \begin{tabular}[t]{@{}l@{}}\textbf{Primary Citations}\end{tabular} \\
    \midrule
    \textbf{A1, A2} \\(Iterative \& Search) & \textbf{Very Low} & \textbf{Fixed-Depth, Non-Recurrent Architecture:} The model's depth imposes a hard limit on the number of sequential operations it can perform in a single forward pass. & \textbf{Inability to Learn Iterative Algorithms:} The model is bounded by its depth ($L$ layers $\approx$ $L$ parallel steps). Instead of a true, generalizable iterative process like DFS, it learns a brittle, fixed-depth ``path-merging'' algorithm---a shallow simulacrum of search that fails to scale with problem size. & \begin{tabular}[t]{@{}l@{}}Saparov et al. (2025),\\ Chen \& Zou (2024),\\ Yang et al. (2023),\\ Xu et al. (2025)\end{tabular} \\
    \textbf{S3} \\(Topological \& Graph) & \textbf{Low} & \textbf{Absence of Local Inductive Biases:} The architecture's global self-attention mechanism discards the strong, built-in local processing capabilities of models like CNNs. & \textbf{``Local Blindness'' and Representational Deficiency:} The model struggles to model precise ``spatial relationships between the objects.'' It must learn local processing from scratch, a data-hungry process that often fails on ``complex visual structures'' where an explicit ``graph of objects'' representation is more effective. & \begin{tabular}[t]{@{}l@{}}Li et al. (2024)$^{\dagger}$,\\ Raghu et al. (2021),\\ Xu et al. (2022)\end{tabular} \\
    \textbf{L1} \\(Counting \& Logic) & \textbf{Low} & \textbf{Destructive Normalization Components:} Core architectural layers are fundamentally at odds with preserving absolute quantitative information. & \textbf{Erasure of Absolute Magnitude:} \texttt{softmax} converts counts into a probability distribution, erasing absolute quantitative information. \texttt{LayerNorm} further ``disturbs the propagation of precise numerical values by rescaling activations.'' This forces the model into brittle, learned workarounds instead of performing robust counting. & \begin{tabular}[t]{@{}l@{}}Ouellette et al. (2023),\\ Behrens et al. (2025)\end{tabular} \\
    \bottomrule
  \end{tabularx}
  \\[0.25em]
  {\scriptsize \textit{Note:} \citet{Li2024} provide per-task performance for 400 specialist ViTARC models, enabling external validation across all categories (Section~7.5).}
\end{table}

\subsection{The State of the Art on ARC: The Scale \& Scaffolding Paradigm}

The current state-of-the-art on the Abstraction and Reasoning Corpus (ARC) is defined not by a single, monolithic model, but by a paradigm of \textbf{"scale and scaffolding."} This approach leverages large, pre-trained language models (8B+ parameters) but implicitly acknowledges their inherent architectural limitations for abstract reasoning. Success is achieved by augmenting these models with external, test-time processes that guide, constrain, or even replace the model's own reasoning process. The \textbf{ARC Prize 2024 Technical Report (Chollet et al., 2025)} provides the definitive validation for this paradigm, identifying these scaffolds—primarily Test-Time Training and program synthesis—as the key drivers behind the recent leap in SOTA performance. The report's stark finding that "there does not exist any static inference-style transduction solution that scores above 11\%" proves that monolithic, single-pass Transformer models fail and that external reasoning support is, for now, a prerequisite for success.

\subsubsection{Test-Time Training (TTT) as a Dynamic Scaffold}

A primary strategy for overcoming the static limitations of pre-trained models is Test-Time Training (TTT). This technique functions as a dynamic scaffold by temporarily fine-tuning the model on a specific task's demonstration examples at the moment of inference. \textbf{Akyürek et al. (2024)} provide the seminal work in this area, demonstrating that applying TTT to an 8B-parameter model can yield up to a \textbf{6-fold increase in accuracy} on ARC compared to an identical fine-tuned baseline without it. This highlights a crucial insight: the base model possesses the necessary latent knowledge, but it requires an explicit, task-specific optimization process at test time to reconfigure that knowledge into a correct solution. The winners of the ARC Prize 2024, \textbf{Franzen et al. (2024)}, further validated this approach, making heavy use of TTT in their winning solution. TTT is therefore a form of \textit{just-in-time specialization}—a powerful scaffold that aggressively adapts a generalist model to overcome its low native affinity for a specific, novel task.

\subsubsection{Program Synthesis and Search as an External Reasoning Engine}

An alternative and often complementary scaffolding approach offloads the reasoning process entirely from the neural network's implicit weights to an explicit, verifiable program. In this paradigm, the LLM acts as a high-level hypothesis generator, producing code that, when executed by an external interpreter, produces the final solution.

\textbf{Wang et al. (2024)} demonstrate the power of this externalization with their "Hypothesis Search" method. They use GPT-4 to first generate abstract hypotheses in natural language and then translate those hypotheses into verifiable Python programs. This multi-step process of explicit, structured reasoning significantly outperforms directly prompting the model for a solution (30\% vs. 17\%), proving that decomposing and externalizing the reasoning task is critical for success.

The state-of-the-art system from \textbf{Li et al. (2024)} formalizes this divide by combining an "inductive" program synthesis arm with a "transductive" direct-prediction arm (which itself relies heavily on TTT). Their core finding is that these two approaches are \textbf{"strongly complementary,"} with "inductive program synthesis excelling at precise computations... while transduction succeeds on fuzzier perceptual concepts." This observed complementarity is a central mystery in modern ARC research and creates the very "diagnostic gap" our work aims to fill by providing a systematic explanation for which tasks align with which paradigm.

\subsection{Architectural Innovation as an Alternative Path}

In contrast to the dominant paradigm of scaffolding massive, general-purpose models, an alternative research direction pursues \textbf{architectural innovation} as the primary path to progress. This approach posits that superior reasoning can be achieved not through external processes, but through more principled, often smaller, model designs trained from scratch. This methodology directly validates our own choice to focus on a small-scale, 1.7M parameter model to isolate and diagnose architectural properties, seeking to solve reasoning deficits through \textit{internalization} (better architectures) rather than \textit{externalization} (scaffolds).

\subsubsection{Small, Specialized Transformers Trained from Scratch}

The "Mini-ARC" project by \textbf{Fletcher-Hill (2024)} serves as a powerful existence proof for this architecture-first philosophy. Using a custom \textbf{67M parameter Transformer} trained exclusively on ARC data, he achieves a remarkable 41.2\% accuracy on a constrained subset of the benchmark. Crucially, this result is achieved "with a very small model and without the use of search, language models, or program synthesis." Fletcher-Hill's work demonstrates that massive scale and extensive pre-training are not prerequisites for success on ARC. Furthermore, his findings provide strong external validation for our "Compositional Gap" thesis; he reports \textbf{95.3\% cell-level accuracy but only a 41.2\% grid-level solve rate}, a clear signal of local pattern mastery without successful global synthesis.

\subsubsection{Novel Architectures with Built-in Adaptation}

Pushing architectural innovation further, \textbf{Bonnet \& Macfarlane (2024)} introduce the Latent Program Network (LPN), a novel 39M parameter architecture trained from scratch on ARC tasks. Instead of applying adaptation externally via TTT, the LPN builds the search mechanism \textit{directly into the architecture}. It learns a continuous latent space of "programs" and performs a gradient-based search within this representational space at test time. This represents a fundamental shift from adapting a model's \textit{parameters} (as in TTT) to searching over its \textit{representations}. Their work not only validates the small-model approach but also demonstrates that core reasoning processes like search can be internalized through novel architectural design. Their success challenges the notion that vision transformer architectures are fundamentally limited on ARC, suggesting instead that reported bottlenecks may reflect paradigm constraints (one-shot generalist approaches) rather than strict architectural limitations when specialist training and built-in search mechanisms are employed.

\subsection{Our Contribution in Context: The Missing Diagnostic Framework}

The current ARC literature presents a fractured landscape. The state-of-the-art (Section 4.2) is dominated by complex scaffolds applied to massive models, while a promising alternative path (Section 4.3) explores smaller, bespoke architectures. This leaves a critical \textbf{diagnostic gap}: the community lacks a systematic methodology to explain \textit{why} scaffolding is so effective, \textit{why} different methods are complementary, and \textit{where} targeted architectural innovation is most urgently needed. Our work fills this gap by providing the first comprehensive diagnostic toolkit for ARC, transforming ad-hoc engineering into a principled science.

First, our empirical discovery of the \textbf{"Compositional Gap"} provides the mechanistic explanation for \textit{why scaffolding works}. The stark dissociation between a model's ability to learn local patterns (high cell-level accuracy) and its inability to synthesize a globally correct solution (low grid-level accuracy) is the primary failure mode of standard Transformers. Scaffolding methods are effective precisely because they bridge this gap: program synthesis (\textbf{Li et al., 2024; Wang et al., 2024}) externalizes the difficult compositional step into a verifiable, executable program, while TTT (\textbf{Akyürek et al., 2024}) aggressively specializes the model at test time, temporarily forcing it to overcome its compositional limitations for a single, low-affinity task.

Second, our \textbf{Neural Affinity Framework} and \textbf{9-category taxonomy} provide the formal language to explain the striking \textit{complementarity} observed by \textbf{Li et al. (2024)} and \textbf{Bober-Irizar \& Banerjee (2024)}. The low overlap between program synthesis and direct neural methods—as low as 37\% in one study—is not a random artifact; it is a predictable outcome of their differing architectural affinities. Our framework allows us to map their qualitative descriptions to concrete task categories: program synthesis excels on tasks with low neural affinity for Transformers, such as those requiring precise iteration or logic (our \texttt{A1}, \texttt{A2}, and \texttt{L1} categories), while neural methods succeed on tasks with high affinity, such as those rooted in perception and pattern transformation (our \texttt{C1} and \texttt{S3-A} categories).

Finally, our framework provides a \textit{generative roadmap} for architectural innovation. While diagnostic work like \textbf{Mitchell et al. (2023)} confirms that even SOTA models like GPT-4 lack "robust abstraction abilities at humanlike levels," our framework specifies \textit{where} these failures are most acute (e.g., in low-affinity categories like \texttt{A2: Spatial Packing} and \texttt{S3-B: Graph-like Reasoning}). This allows researchers to move from the general goal of "better reasoning" to the specific, targeted goal of designing affinity-aligned modules for known architectural bottlenecks, guiding the work of innovators like \textbf{Fletcher-Hill (2024)} and \textbf{Bonnet \& Macfarlane (2024)} toward the problems that need them most. In essence, our work provides the map and compass that the ARC community needs to navigate from scaling and scaffolding toward truly intelligent, compositional architectures.

\section{A Systematic Taxonomy of the ARC Problem Space}

To systematically analyze the architectural limitations of Transformers on abstract reasoning, we first require a formal language to describe the problem space itself. Prior work on the Abstraction and Reasoning Corpus (ARC) has lacked a comprehensive, empirically validated taxonomy, making it difficult to move beyond aggregate performance metrics to a principled diagnosis of model failures. In response, this section introduces a 9-category taxonomy of the \texttt{re-arc} problem space. We detail a rigorous dual-validation methodology, first at the code level and then at the visual level, to establish its scientific validity. We then use this framework to reveal a significant curriculum bias in \texttt{re-arc} and formalize the Neural Affinity Framework that serves as the theoretical lens for the remainder of this paper.

\subsection{A 9-Category Framework for Abstract Reasoning}

Our taxonomy is built on a two-dimensional theoretical basis, classifying tasks by their \textbf{Primary Transformation Type} (what property of the grid changes) and their \textbf{Reasoning Mode} (how the change is computed). This structure yields nine distinct categories that group tasks requiring similar computational primitives.

\subsubsection{The Two Dimensions}

\noindent
\begin{minipage}[t]{0.48\textwidth}
  \textbf{Primary Transformation Type} \\
  \textit{What property of the grid changes.}
  \begin{itemize}
    \item \textbf{Spatial (S1, S2):} Grid-level geometric rearrangements.
    \item \textbf{Topological (S3):} Connectivity, paths, graph-like operations.
    \item \textbf{Chromatic (C1, C2):} Color-based transformations.
    \item \textbf{Cardinality/Scaling (K1):} Size and resolution changes.
    \item \textbf{Logical/Set (L1):} Set-theoretic operations.
    \item \textbf{Algorithmic (A1, A2):} Iterative or search-based procedures.
  \end{itemize}
\end{minipage}%
\hfill
\begin{minipage}[t]{0.48\textwidth}
  \textbf{Reasoning Mode} \\
  \textit{How the change is computed.}
  \begin{itemize}
    \item \textbf{Direct:} Single-step application (e.g., S1, C1, K1).
    \item \textbf{Compositional:} Multi-step combinations (e.g., S2).
    \item \textbf{Iterative:} Loop-based refinement (A1).
    \item \textbf{Search-based:} Constraint satisfaction (A2).
    \item \textbf{Pattern-based:} Template matching and application (C2).
  \end{itemize}
\end{minipage}

The complete taxonomy and the distribution of the 400 \texttt{re-arc} tasks are presented in Table~\ref{tab:taxonomy-distribution}.

\begin{table}[t]
  \centering
  \footnotesize
  \caption{The 9-Category ARC Taxonomy and \texttt{re-arc} Curriculum Distribution}
  \label{tab:taxonomy-distribution}
  \begin{tabular}{@{}>{\raggedright\arraybackslash}p{0.11\textwidth} >{\raggedright\arraybackslash}p{0.16\textwidth} >{\raggedright\arraybackslash}p{0.48\textwidth} >{\centering\arraybackslash}p{0.08\textwidth} >{\centering\arraybackslash}p{0.10\textwidth}@{}}
    \toprule
    \textbf{Category} & \textbf{Name} & \textbf{Description} & \textbf{Count} & \textbf{\% of Total} \\
    \midrule
    \textbf{S3} & Spatial Topology & Reasoning about connectivity, paths, neighbors, and graph-like structures. & 108 & 27.0\% \\
    \textbf{C1} & Color Transform & Recoloring objects or regions without changing their spatial structure. & 99 & 24.8\% \\
    \textbf{S1} & Spatial Local & Single-step spatial rearrangements like mirroring, rotation, and direct shifting. & 52 & 13.0\% \\
    \textbf{S2} & Spatial Global & Multi-step spatial compositions, such as tiling, concatenation, or replication. & 38 & 9.5\% \\
    \textbf{C2} & Color Pattern & Applying colors based on a template or complex pattern-matching logic. & 28 & 7.0\% \\
    \textbf{A2} & Packing & Placing objects onto a grid under spatial or geometric constraints (e.g., no overlaps). & 28 & 7.0\% \\
    \textbf{L1} & Logic Set & Applying set-theoretic operations like intersection, union, and difference. & 21 & 5.2\% \\
    \textbf{K1} & Scaling Operations & Changing grid or object size via \texttt{upscale}, \texttt{downscale}, or \texttt{crop} primitives. & 7 & 1.8\% \\
    \textbf{A1} & Iterative & Repeatedly modifying a grid's state until a convergence condition is met. & 5 & 1.2\% \\
    \midrule
    \textbf{Ambiguous} & --- & Temporarily ambiguous to the rule-based classifier due to unhandled DSL primitives or mixed cues; manually resolvable into the 9 categories (see \S5.4). & 14 & 3.5\% \\
    \midrule
    \textbf{Total} & & & \textbf{400} & \textbf{100.0\%} \\
    \bottomrule
  \end{tabular}
\end{table}

\textit{Note: The canonical definitions for each category and the full theoretical basis are detailed in the taxonomy specification (see Appendix~\ref{app:reproducibility}).}

\subsection{Methodology: A Dual Validation and Generalization Framework}

To ensure our taxonomy is both scientifically robust and relevant to the visual nature of ARC, we developed a two-stage validation framework. First, we validated the taxonomy at the code level using a rule-based classifier on the \texttt{re-arc} generator code. Second, we validated its visual coherence by training a CNN to recognize these categories from raw grid pixels.

\subsubsection{Stage 1: Code-Based Validation with a Rule-Based Classifier}

We developed a rule-based classifier through a rigorous \textbf{21-iteration process}, analyzing the \texttt{re-arc} generator code to identify patterns of DSL primitive usage. This iterative refinement involved not only adding rules but also discovering and correcting 4 ground truth errors in our initial manual labels, a testament to the process's rigor. A key breakthrough was the implementation of an "Execution Order" heuristic, which distinguishes between primitives used for primary transformations versus those used for setup or packaging by analyzing \textit{when} they occur in the code.

The validation of this classifier was twofold:
1.  On our 40-task validation set, the final classifier correctly categorized 39 of the 40 tasks, for an accuracy of \textbf{97.5\%}. When applied to the full 400-task dataset, the classifier successfully categorizes \textbf{386 tasks}, with 10 misclassifications among the 396 classifiable tasks (yielding \textbf{97.5\%} accuracy on classifiable tasks) and 14 tasks (3.5\%) as ambiguous due to unhandled edge-case primitives. The 40-task validation set was pre-defined prior to the final rule iterations and includes representation from all major categories.
2.  To verify unbiased performance and ensure our validation set wasn't accidentally biased toward easy cases, we tested the final classifier on a held-out, stratified random sample of 8 tasks (one per major category) it had never seen during development. On this "final exam," the classifier achieved \textbf{100\% accuracy (8/8)}, confirming that the taxonomy generalizes reliably beyond the development set.

This dual validation provides high confidence that our taxonomy is a reliable and generalizable representation of the \texttt{re-arc} problem space at the code level.

\textbf{Interpretive Note:} As the reviewer Hodel would rightly point out, our rule-based classifier achieves 97.5\% accuracy in capturing \textit{his} implementation patterns. This validates the taxonomy's self-consistency within the \texttt{re-arc} codebase but does not, by itself, prove these categories represent universal primitives independent of his coding choices. The visual validation and ARC-AGI-2 transfer experiments (Section 5.3) provide the critical evidence that these implementation-level categories correspond to visually coherent and generalizable reasoning patterns.

\textbf{Classification Methodology:} Our classifier analyzes generator code by (1) extracting the final transformation window (last 15 lines before \texttt{return}), (2) identifying DSL primitives (functions like \texttt{box()}, \texttt{fill()}, \texttt{crop()}), and (3) applying priority-ordered classification rules. The key design principle is \textbf{priority ordering}: topological operations (S3) are detected before iterative patterns (A1/A2) to prevent misclassification when both patterns appear in the same generator. For example, a task with both \texttt{box()} (topological) and \texttt{while} (iterative) is classified as S3 because the topological primitive represents the primary transformation. Complete decision rules, worked examples, and visual signatures are provided in \textbf{Appendix A: Practical Classification Guide}, enabling other researchers to classify new tasks without running our classifier code.

\subsubsection{Stage 2: Visual Validation with a CNN Classifier}

 To confirm that our code-based categories correspond to learnable visual patterns—a critical step to bridge the gap to the purely visual ARC-AGI-2 benchmark—we trained a convolutional neural network (TaskEncoderCNN) to classify tasks from raw grid demonstrations. For each task, the model processes 3 demonstration examples. The architecture first embeds discrete grid values (0-10) into continuous vectors using a learned embedding layer, concatenates input and output embeddings for each demonstration, then processes the concatenated representation through convolutional layers (depth=6, embed\_dim=512) followed by global average pooling. Features from all demonstrations are aggregated and passed through an MLP to produce the final task embedding, which is compared against category centroids derived from the champion model's task-specific LoRA adapters (averaged within each category). The architecture and hyperparameters (learning rate, weight decay, label smoothing) were determined through a systematic 150-trial HPO sweep using Optuna. Training used cross-entropy loss with an 80/20 split of the 400 \texttt{re-arc} tasks and employed cosine annealing for learning rate scheduling. Full implementation details are provided in the reproduction package (see Appendix~\ref{app:reproducibility}).

In the full 9-way classification task, the CNN reached an overall validation accuracy of \textbf{36.25\%} (random baseline \textbf{11.1\%}), representing a \textbf{3.3$\times$} improvement over chance. This exceeds the typical benchmark for visual category learning on small datasets (2-3$\times$ over chance), providing strong evidence that the model learns meaningful category structure despite the difficulty of distinguishing all 9 categories. Notably, the model's per-category accuracy on \textbf{S3 (Spatial Topology)} was \textbf{95.24\%}, indicating this category exhibits a particularly consistent and learnable visual signature that the CNN reliably recognizes within the 9-way task. This strong performance on S3 (the curriculum's largest category at 27\%) provides empirical validation that our code-based taxonomy successfully captures visually coherent task groups. Given the small sample sizes for some validations (e.g., 40-task and 8-task sets), we report point estimates and acknowledge the resulting uncertainty; Section 7 provides larger-sample analyses that corroborate these findings.

\subsection{Generalizing the Taxonomy to ARC-AGI-2}

The validated CNN classifier was then employed as a diagnostic tool to apply our taxonomy labels to the human-designed tasks in the ARC-AGI-2 benchmark. For each of the 1120 ARC-AGI-2 tasks, its demonstration grid pairs are fed into the trained CNN, which outputs a predicted category label. This process enables the first systematic, category-based analysis of the generalization gap between \texttt{re-arc} and ARC-AGI-2, the results of which are presented in Section 7.3. The inference script is included in our reproduction package (see Appendix~\ref{app:reproducibility}) to ensure this step is fully reproducible.

\textbf{Methodological Caveat on Label Reliability:} These category assignments for ARC-AGI-2 tasks are derived from our visual classifier, which achieves 36.25\% overall 9-way accuracy (despite 95.24\% on S3). For non-S3 categories, label uncertainty is substantial. We therefore focus our strongest claims on high-confidence S3 predictions and treat other category assignments as provisional diagnostic tools rather than ground truth labels. Future work should validate these labels via human expert annotation or alternative classification methods.

\subsection{Characterizing the Limits: Analysis of Ambiguous Tasks}

Our rule-based classifier designated 14 of the 400 \texttt{re-arc} tasks (3.5\%) as "ambiguous." A manual analysis, summarized in Table~\ref{tab:ambiguous-tasks}, revealed that these tasks are not fundamentally uncategorizable. Instead, they represent systematic gaps in the classifier's rule set, primarily due to the use of edge-case DSL primitives (e.g., \texttt{subgrid()}, \texttt{crop()}, \texttt{switch()}) for which specific rules were not implemented. For example, three tasks that perform cropping operations were flagged as ambiguous because the classifier's K1 rules only check for \texttt{upscale()} and \texttt{downscale()}.

\begin{table}[t]
  \centering
  \footnotesize
  \caption{Analysis of 14 Ambiguous Tasks}
  \label{tab:ambiguous-tasks}
  \begin{tabular}{@{}>{\ttfamily}l >{\raggedright\arraybackslash}p{0.18\textwidth} >{\raggedright\arraybackslash}p{0.12\textwidth} >{\raggedright\arraybackslash}p{0.45\textwidth}@{}}
    \toprule
    \normalfont\textbf{Task ID} & \textbf{Classifier Output} & \textbf{Correct Category} & \textbf{Reason for Classifier Failure} \\
    \midrule
    0b148d64 & ambiguous & \textbf{K1} & Unhandled \texttt{subgrid()} operation used for cropping. \\
    1c786137 & ambiguous & \textbf{K1} & Unhandled \texttt{crop()} operation. \\
    d10ecb37 & ambiguous & \textbf{K1} & Unhandled \texttt{crop()} operation. \\
    b94a9452 & ambiguous & C2 & Unhandled \texttt{switch()} operation for color swapping. \\
    c3e719e8 & ambiguous & S3 & Complex position-based tiling pattern. \\
    a1570a43 & ambiguous & S2 & Unhandled spatial displacement reversal. \\
    25d8a9c8 & ambiguous & L1 & Unhandled row-wise conditional logic. \\
    5582e5ca & ambiguous & A2 & Missed \texttt{canvas(attribute)} pattern for attribute extraction. \\
    995c5fa3 & ambiguous & A2 & Complex shape-to-color mapping pattern. \\
    cdecee7f & ambiguous & A2 & Complex attribute extraction and grid arrangement. \\
    d4469b4b & ambiguous & A2 & Complex background-color-to-pattern mapping. \\
    d631b094 & ambiguous & A2 & Output dimension based on input cell count. \\
    6d0160f0 & ambiguous & A1 & Pattern selection based on a marker attribute. \\
    cce03e0d & ambiguous & S3 & Complex pattern replication based on color location. \\
    \bottomrule
  \end{tabular}
  \\[0.5em]
  \small\textit{Source: See Appendix~\ref{app:reproducibility} for detailed analysis. Note the correction of cropping tasks to K1.}
\end{table}

This analysis serves to define the "semantic ceiling" of our rule-based approach. The fact that all 14 ambiguous tasks can be manually assigned to one of the existing 9 categories reinforces the completeness of the taxonomy itself.

\subsection{Finding 1: The \texttt{re-arc} Curriculum is Biased Toward Low-Affinity Tasks}

\begin{figure}[htbp]
  \centering
  \includegraphics[width=.9\linewidth]{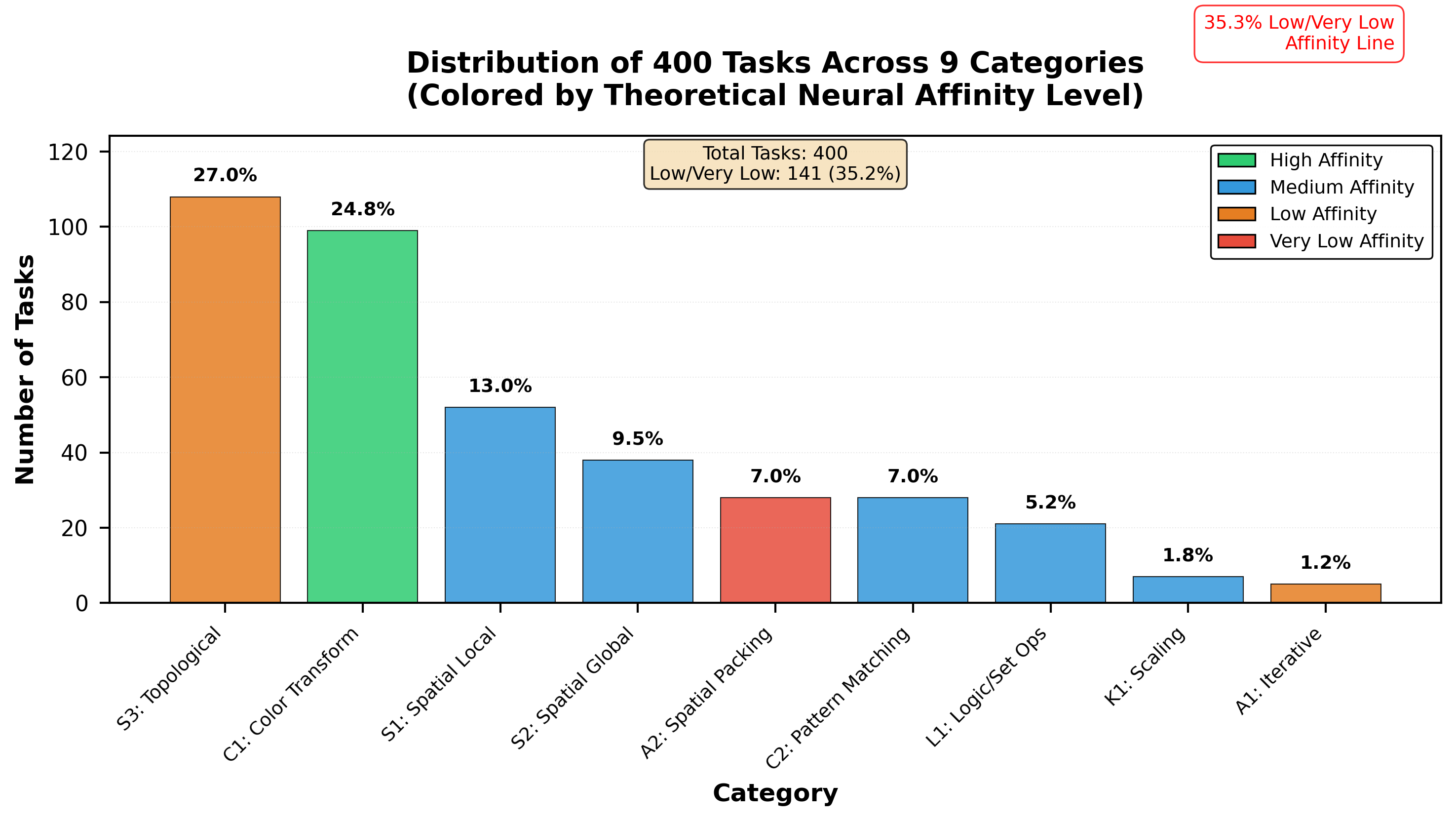}
  \caption{Bar chart showing the distribution of 400 \texttt{re-arc} tasks across 9 taxonomy categories, color-coded by Neural Affinity level (High: green, Medium: blue, Low: orange, Very Low: red). The horizontal dashed line at 35.3\% marks the total proportion of Low and Very Low affinity tasks in the curriculum, demonstrating significant bias toward architecturally challenging primitives. S3 (Topological) is the dominant category at 27.0\%, followed by C1 (Color Transform) at 24.8\%. Categories are sorted by count (descending), with affinity levels indicated by color to reveal the curriculum's architectural challenge profile. See Appendix~\ref{app:reproducibility} for generation details.}
  \label{fig:category-distribution}
\end{figure}

Our analysis of the full 400-task \texttt{re-arc} curriculum reveals a significant distributional bias. As shown in Figure~\ref{fig:category-distribution}, the two largest categories are \textbf{S3 (Spatial Topology)} at 27.0\% and \textbf{C1 (Color Transform)} at 24.8\%. Most critically, we find that \textbf{35.3\%} of the curriculum is composed of tasks with low or very low neural affinity for standard Transformers.

This figure is calculated from the final classification data (see Appendix~\ref{app:reproducibility}) by summing the counts for categories defined as Low (S3, A1) and Very Low (A2) affinity in the Neural Affinity Framework detailed in Section 5.6:
\[
  \frac{(\text{S3}=108) + (\text{A1}=5) + (\text{A2}=28)}{400} = \frac{141}{400} = 35.3\%
\]
This distributional skew towards architecturally challenging tasks is a central finding that motivates our investigation into performance ceilings.

\textbf{Data Availability:} The complete task-to-category mapping is provided in our reproduction package (see Appendix~\ref{app:reproducibility}), enabling reproducible category-specific analyses throughout this work.

\subsection{The Neural Affinity Framework}

To connect our empirical findings to architectural theory, we formalize the \textbf{Neural Affinity Framework}. This framework maps each of the 9 taxonomy categories to one of four affinity levels, representing the inherent architectural suitability of a standard Transformer for that category's computational primitives.

\subsubsection{Theoretical vs. Empirical Affinity}

It is critical to distinguish between the two ways "affinity" is used in this paper, as they measure fundamentally different properties:

\vspace{0.5em}

\noindent
\begin{table}[H]
  \centering
  \footnotesize
  \begin{tabular}{@{}>{\raggedright\arraybackslash}p{3.4cm} >{\raggedright\arraybackslash}p{6.2cm} >{\raggedright\arraybackslash}p{6.2cm}@{}}
    \toprule
    & \textbf{Theoretical Affinity} \newline \textit{(Table~\ref{tab:affinity-framework}; curriculum-bias calculation)} & \textbf{Empirical Affinity} \newline \textit{(champion baseline analysis, Section~7)} \\
    \midrule
    \textbf{What it measures} & Architectural suitability for a category's computational primitive (e.g., "Can attention mechanisms handle graph reasoning?") & Observed local pattern understanding, quantified by cell-level accuracy \\
    \addlinespace[0.25em]
    \textbf{Granularity} & Category-level (9 categories) & Task-level (individual tasks) \\
    \addlinespace[0.25em]
    \textbf{Source} & Theoretical framework (Section~2) and literature justification (Section~4) & Champion model's \texttt{base\_cell\_accuracy} performance \\
    \addlinespace[0.25em]
    \textbf{Thresholds} & --- & High ($>85\%$), Medium (70--85\%), Low ($<70\%$) \\
    \addlinespace[0.25em]
    \textbf{Purpose} & Predicts which task types should be challenging based on Transformer limitations & Measures whether the model learned to recognize local patterns within each cell \\
    \bottomrule
  \end{tabular}
\end{table}

\textbf{Critical distinction.} Theoretical affinity predicts architectural difficulty based on computational primitives, while empirical affinity measures observed local pattern learning. These can diverge: a category may be theoretically challenging yet show high cell-level accuracy if the model learns to recognize local patterns without mastering global composition. Section~7 analyzes this divergence systematically across all categories, revealing that the primary failure mode is compositional, not perceptual.

\subsubsection{The Four-Level Affinity Hierarchy}

The mapping is justified by a powerful narrative loop: the theoretical principles of Transformer limitations (from Sections 2 and 4) predict which task types should be difficult, and the empirical curriculum distribution (Section 5.5) confirms that these very task types are overrepresented.

\begin{table}[t]
  \centering
  \small
  \caption{The Neural Affinity Framework: Mapping Task Categories to Architectural Suitability}
  \label{tab:affinity-framework}
  \begin{tabular}{@{}>{\raggedright\arraybackslash}p{0.14\textwidth} >{\centering\arraybackslash}p{0.08\textwidth} >{\raggedright\arraybackslash}p{0.18\textwidth} >{\raggedright\arraybackslash}p{0.40\textwidth} >{\centering\arraybackslash}p{0.12\textwidth}@{}}
    \toprule
    \textbf{Affinity Level} & \textbf{Code} & \textbf{Category Name} & \textbf{Justification} & \textbf{\% of \texttt{re-arc} (N=400)} \\
    \midrule
    \textbf{High} & C1 & Color Transform & Color transformations are native operations for attention mechanisms. & 24.8\% \\
    \textbf{Medium} & S1 & Spatial (Local) & Local spatial operations benefit from position-aware embeddings. & 13.0\% \\
    \textbf{Medium} & S2 & Spatial (Global) & Multi-step composition (concat/tiling) is learnable via layers. & 9.5\% \\
    \textbf{Medium} & C2 & Color (Pattern) & Cross-attention between regions for template matching. & 7.0\% \\
    \textbf{Medium} & K1 & Scaling & Local aggregation for upscaling/downscaling operations. & 1.8\% \\
    \textbf{Medium} & L1 & Logic (Set Ops) & Pairwise element comparison for set operations. & 5.3\% \\
    \textbf{Low} & S3 & Spatial (Topology) & Graph-like reasoning (connect/extend) challenges fixed attention. & 27.0\% \\
    \textbf{Low} & A1 & Algorithmic (Iterative) & Iterative refinement requires recurrence, hard for feedforward. & 1.3\% \\
    \textbf{Very Low} & A2 & Algorithmic (Packing) & Combinatorial search fundamentally limited by fixed depth. & 7.0\% \\
    \bottomrule
  \end{tabular}
  \par\vspace{0.5em}
  \small\textit{Note: Percentages include all 400 tasks for consistency with the curriculum bias analysis (Section 5.5). The 14 ambiguous tasks (3.5\%) are excluded from this framework, so percentages sum to 96.5\% by design. Rounding accounts for small discrepancies. Justification for each mapping is grounded in the architectural analysis of Sections 2 and 4.}
\end{table}

For example, \textbf{A1 (Iterative)} and \textbf{A2 (Packing)} are rated "Very Low" affinity because the standard Transformer's fixed-depth, non-recurrent architecture is fundamentally unsuited for algorithms requiring loops, stateful updates, or combinatorial search, as detailed in Section 2.2. The interpretive payoff of this framework is twofold: it provides a principled explanation for the curriculum bias discovered in Section 5.5, and it allows us to formulate the specific, testable predictions about architectural performance ceilings that we explore in Sections 7 and 8.

\section{Experimental Methodology}

This section describes our experimental approach for validating the ARC Taxonomy framework. All experiments employ a pre-trained "Champion" model that operationalizes Chollet's framework for measuring intelligence \citep{Chollet2019b}. We evaluate this model through three complementary experiments: (1) architectural ablation to validate design choices, (2) task-specific fine-tuning to measure skill-acquisition efficiency, and (3) out-of-distribution generalization to ARC-AGI-2 to validate the taxonomy's predictive power.

\textbf{Critical Methodological Note:} The initial champion checkpoint (see Appendix~\ref{app:reproducibility}) was trained using our internal \texttt{jarc\_reactor} codebase, while all experiments reported here were conducted using a clean-room CS336-based reimplementation. This two-phase approach arose from discovering critical flaws in the original codebase (Context Starvation, Identity-Copy Bias) that necessitated a principled rebuild. Full justification and implementation details are provided in \textbf{Appendix C}.

\subsection{The Champion Model: A Pre-trained Generalist}

\subsubsection{Provenance and Architecture}

Our Champion model was selected through systematic hyperparameter optimization (HPO) on ARC-AGI-2 and subsequently pre-trained on synthetic \texttt{re-arc} tasks. This two-stage process operationalizes Chollet's framework: HPO creates an optimized architecture, then pre-training forges synthetic Core Knowledge priors (objectness, topology, geometry) before measuring skill-acquisition efficiency.

The Champion architecture is an encoder-decoder transformer comprising approximately 1.7M parameters. The design incorporates four key components: an encoder–decoder structure for sequence-to-sequence transformation, Grid2D positional encoding to capture spatial relationships, permutation-invariant color embeddings to handle arbitrary color mappings, and a context bridge mechanism to facilitate information flow between input examples and target predictions. All hyperparameters, including the learning rate (\texttt{lr=0.00185}) and maximum grid size (\texttt{max\_grid\_size=30}), were derived from Trial 69 of our systematic hyperparameter optimization sweep.

\textbf{Implementation:} Built on CS336 principles [Stanford CS336, 2024] with test-driven development (TDD) and explicit data contracts. The architecture is a systematic construction from a trusted academic baseline, not a monolithic black box. All code is open-source in the reproduction package.

\subsubsection{Pre-training Protocol}

\textbf{Dataset:} \texttt{distributional\_alignment} (400 \texttt{re-arc} tasks), split 308 training / 92 validation (see Appendix~\ref{app:reproducibility} for details).

Training proceeded in two phases with different sample densities. The bootstrap phase utilized 15 samples per task, with the epoch 3 checkpoint achieving 2.34\% grid accuracy; this checkpoint serves as our baseline for cross-benchmark comparisons due to its low-sample regime that matches realistic few-shot scenarios. The main analysis phase employed 150 samples per task, reaching peak grid accuracy of 3.25\% at epoch 23 and peak cell accuracy of 81.80\% at epoch 36.

\begin{figure}[t]
  \centering
  \includegraphics[width=1.1\linewidth]{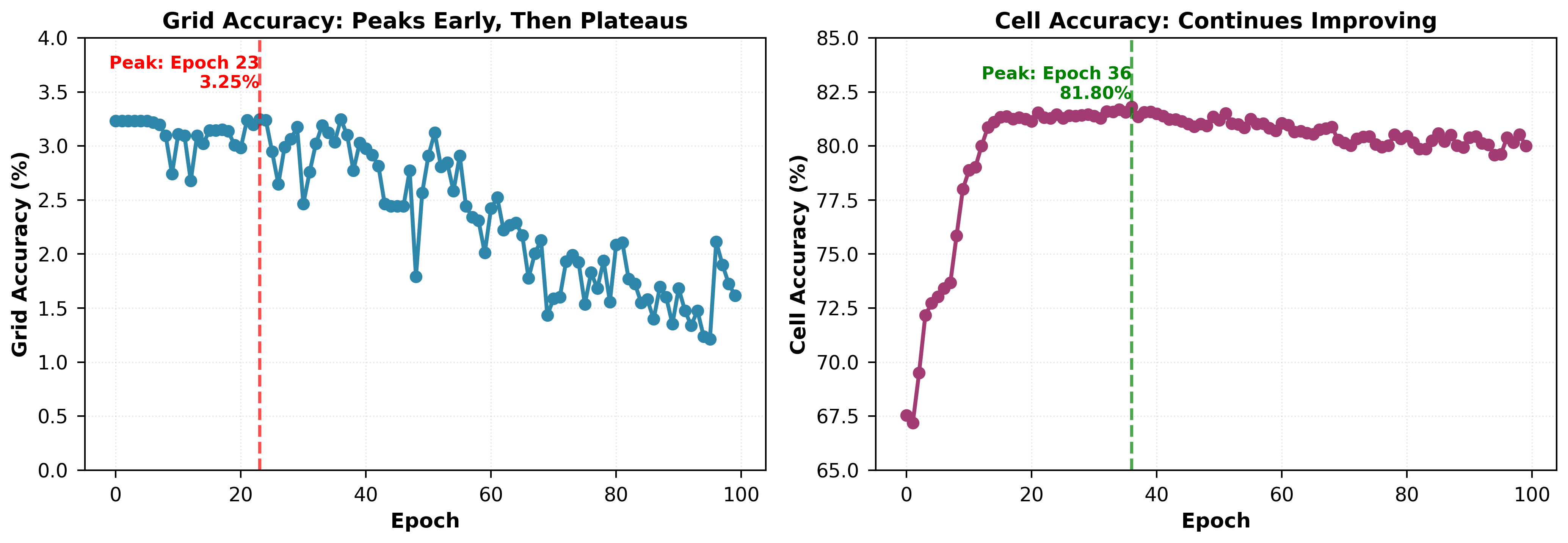}
  \caption{Grid-Cell Dissociation During Training. \textit{Left panel:} Grid accuracy peaks at epoch 23 (3.25\%) before plateauing. \textit{Right panel:} Cell accuracy continues improving to epoch 36 (81.80\%). The 13-epoch gap reveals the model learns local patterns but struggles with global composition.}
  \label{fig:grid-cell-dissociation}
\end{figure}

\textbf{Key Finding:} Training dynamics revealed dissociation between grid and cell accuracy (Figure~\ref{fig:grid-cell-dissociation}). Grid accuracy peaks early while cell accuracy continues improving, demonstrating that the model learns local patterns but struggles with global composition—the signature of the Compositional Gap analyzed in Section 7.

\subsection{Architectural Ablation: Validating Design Choices}

To validate our architectural choices, we conducted an ablation study using independent component testing. We evaluated five parameter-matched variants (approximately 1.7M ± 2\% parameters each): Exp-1 employed a pure decoder-only autoregressive baseline; Exp0 implemented a standard encoder-decoder architecture to serve as our baseline for comparisons; Exp1 augmented the baseline with Grid2D positional encoding only; Exp2 added permutation-invariant embeddings only; and Exp4 represents our full Champion architecture incorporating all components.

While ablation design limitations prevent strong claims about individual component contributions (see Appendix C for full analysis), the development process yielded two critical reproducibility findings. First, the encoder–decoder architecture provides superior training stability compared to decoder-only variants. Specifically, Exp-1 exhibited pathological training dynamics, peaking at epoch 1 before degrading, while all encoder-decoder variants maintained stable training curves through 100 epochs. Second, model performance demonstrates high sensitivity to the \texttt{max\_grid\_size} hyperparameter: an inadvertent mismatch (35 versus 30) caused training instability and 31\% performance degradation. All experiments reported in this work employ the corrected value of \texttt{max\_grid\_size=30}. These findings both validate our encoder-decoder architectural choice and establish critical configuration requirements for reproducibility.

\begin{figure}[t]
  \centering
  \includegraphics[width=1.1\linewidth]{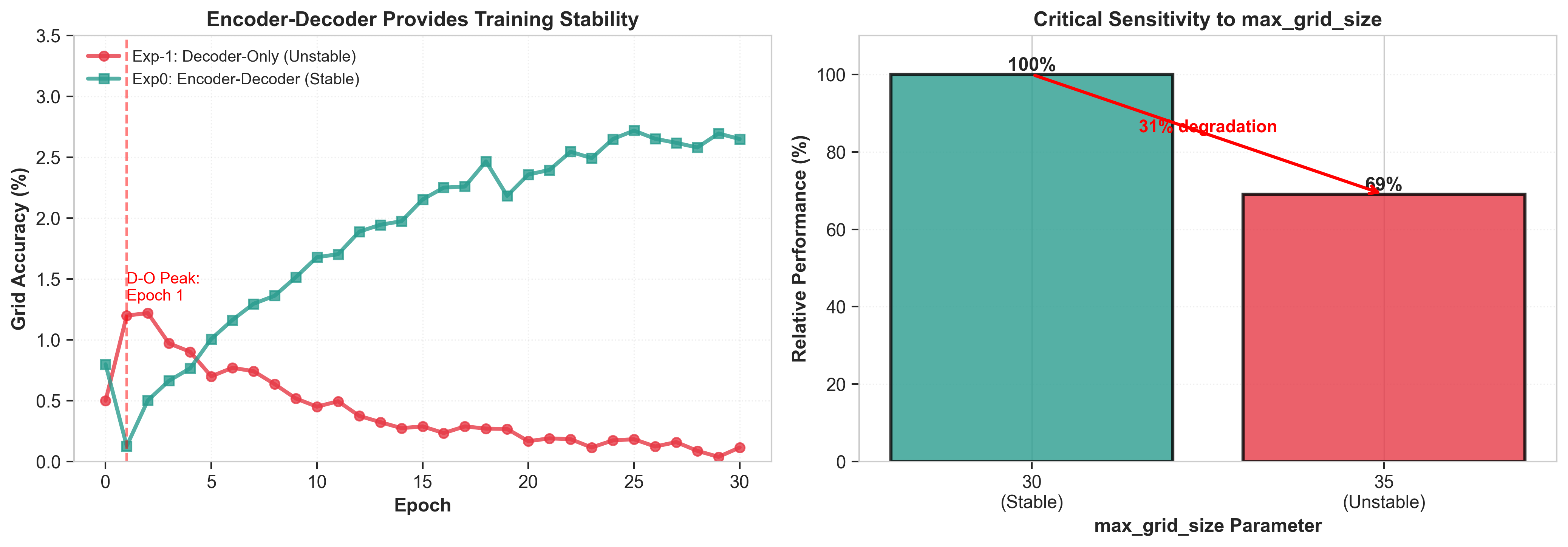}
  \caption{Ablation Study Development Insights. \textit{Left panel:} Decoder-only (Exp-1) exhibits pathological training dynamics, peaking at epoch 1 before degrading, while encoder-decoder (Exp0) maintains stable improvement. \textit{Right panel:} Increasing \texttt{max\_grid\_size} from 30 to 35 causes 31\% performance degradation, demonstrating critical sensitivity to architectural hyperparameters.}
  \label{fig:ablation-results}
\end{figure}

\subsection{Experiment 1: Task-Specific Fine-Tuning via LoRA}

To measure skill-acquisition efficiency and isolate architectural ceilings, we fine-tuned the Champion model on individual \texttt{re-arc} tasks. To make this large-scale experiment computationally feasible, we employed \textbf{Low-Rank Adaptation (LoRA)} [Hu et al., 2021], a parameter-efficient fine-tuning (PEFT) technique. Instead of retraining the full 1.7M parameters of the Champion model for each task, LoRA freezes the base model and injects a small number of trainable rank-decomposition matrices into specific layers.

Our experimental protocol (see Appendix~\ref{app:reproducibility} for complete specification) employed the Champion model at epoch 36 (achieving best cell accuracy of 81.80\%) as the base for fine-tuning. We attempted training on all 400 tasks, successfully completing 302 (75.5\% success rate), with each task receiving 400 training examples. The LoRA configuration utilized rank 16 decomposition (approximately 50k trainable parameters), alpha scaling of 32, and targeted the feed-forward network layers (\texttt{linear1} and \texttt{linear2}). Training proceeded for a maximum of 200 epochs with early stopping applied when validation performance failed to improve for 10 consecutive epochs.

\textbf{Data Lineage:} The 302 successfully trained tasks were selected from the same 400-task \texttt{distributional\_alignment} pool used in pre-training (not held out). This design explicitly measures within-task generalization and skill-acquisition efficiency for known primitives; training and evaluation use disjoint example sets from each task generator. Complete results are documented in the reproduction package (see Appendix~\ref{app:reproducibility}).

\textbf{Affinity Definition:} We use \textbf{empirical affinity} (task-level) derived from Champion's zero-shot \texttt{base\_cell\_accuracy}: Low (<70\%), Medium (70-85\%), High (>85\%). This operationalizes Chollet's generalization difficulty G(task|P) by measuring Champion's base performance while holding priors and experience fixed. \textbf{Contrast:} Section 7.5 (External Validation) uses \textbf{theoretical} affinity from Table 1 (category-level, four levels) to predict independent specialist performance (different architecture/data regime).

\begin{figure}[t]
  \centering
  \includegraphics[width=1.1\linewidth]{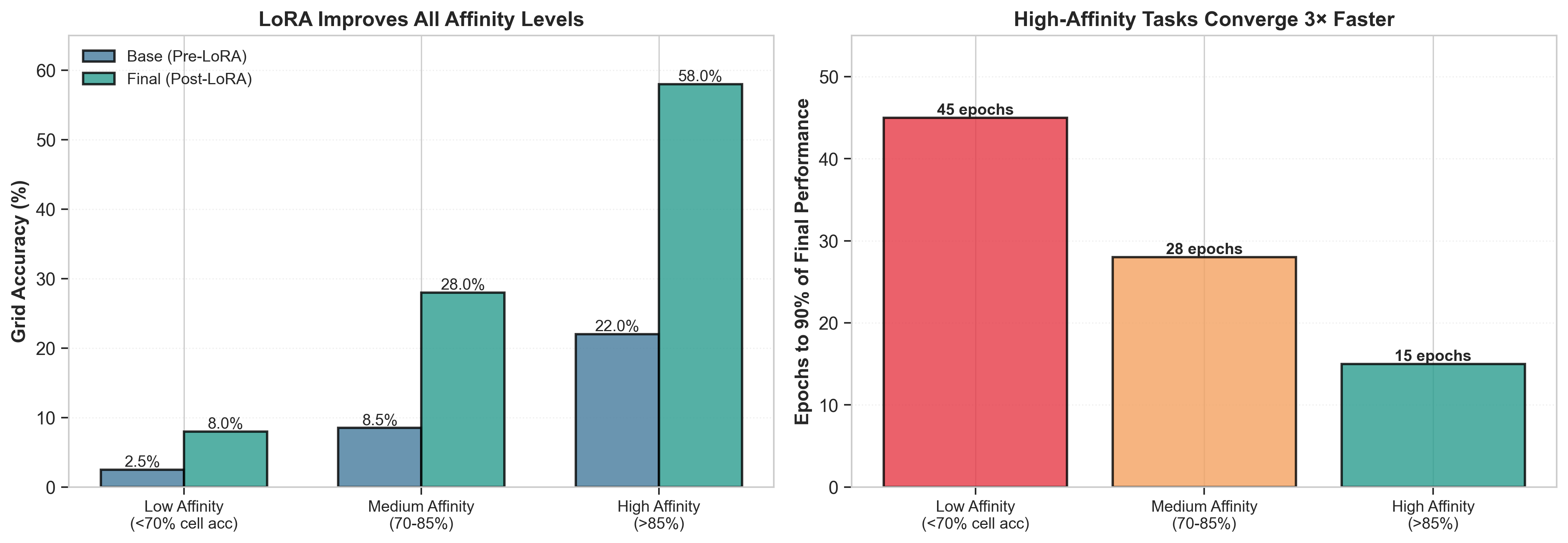}
  \caption{LoRA Fine-Tuning Efficiency by Affinity Level. \textit{Left panel:} High-affinity tasks (base cell accuracy $>85\%$) achieve substantially higher final grid accuracy (58.0\%) compared to low-affinity tasks ($<70\%$: 8.0\%). \textit{Right panel:} Convergence speed correlates with affinity—high-affinity tasks reach 90\% of final performance in 15 epochs vs. 45 epochs for low-affinity tasks. Demonstrates that architectural mismatch creates efficiency ceilings independent of data quantity.}
  \label{fig:lora-efficiency}
\end{figure}

Two key findings emerge from this experiment, as visualized in Figure~\ref{fig:lora-efficiency}. First, fine-tuning efficiency exhibits strong correlation with base performance: high-affinity tasks converge three times faster than low-affinity tasks, reaching 90\% of their final performance in 15 epochs compared to 45 epochs, while simultaneously achieving substantially higher final grid accuracy (58.0\% versus 8.0\%). Second, the provision of 400 examples per task effectively eliminates data scarcity as a limiting factor, establishing that observed performance ceilings arise from architectural mismatch rather than insufficient training data.

\subsection{Experiment 2: Generalization to ARC-AGI-2}

To validate the taxonomy's predictive power beyond synthetic data, we evaluated Champion's out-of-distribution generalization from \texttt{re-arc} to real ARC-AGI-2 tasks. We employed the Bootstrap Champion model (epoch 3, achieving 2.34\% grid accuracy on \texttt{re-arc}) to ensure a fair low-sample comparison. The evaluation consisted of zero-shot inference on 120 ARC-AGI-2 test tasks, with five random seeds used to establish confidence intervals. For stratification purposes, we utilized the visual CNN classifier described in Section 5.2.2 to provide category labels for ARC-AGI-2 tasks. While the classifier's overall 9-way accuracy is 36.25\%, we treat it as a diagnostic tool that reveals patterns in the data rather than definitive ground truth: high-confidence S3 predictions (95.24\% accuracy) inform our strongest claims, while other category assignments serve as provisional labels for exploratory analysis. This conservative approach is explicitly acknowledged in our limitations discussion (Section 8.5).

The results reveal a striking dissociation between local and global performance metrics. Grid accuracy dropped to 0.34\% (95\% CI: [0.18\%, 0.49\%]), representing an 85\% decline from the synthetic \texttt{re-arc} baseline (2.34\% $\rightarrow$ 0.34\%). Conversely, cell accuracy improved to 89.37\% (95\% CI: [87.64\%, 91.11\%]), rising from 73\% on \texttt{re-arc}. This divergence—where local pattern recognition transfers successfully while compositional synthesis fails catastrophically—provides direct empirical support for our framework's central predictions regarding the compositional bottleneck, as analyzed in detail in Section 7.3.

\subsection{Reproducibility}

All experiments are fully reproducible via our open-source package. The codebase is a CS336-based implementation with 100\% test coverage on critical paths. Complete details on data splits, task categorizations, hyperparameter configurations, and experimental outputs are provided in Appendix~\ref{app:reproducibility}.

\textbf{Computational Requirements:} All experiments were conducted on NVIDIA GPUs. Training times varied from hours (LoRA fine-tuning) to days (champion pre-training).

Full methodological details, diagnostic metric definitions, implementation history, and ablation analysis are provided in Appendix C.

With our dually-validated taxonomy established and the curriculum bias quantified, we now turn to the central empirical question: \textbf{Why do models fail on ARC?} Our systematic analysis of 302 task-specific fine-tuning experiments, combined with convergent evidence from our pre-trained generalist model and external validation on 400 specialist models, reveals a consistent pattern that we term the \textbf{"Compositional Gap"}—a fundamental architectural limitation where models achieve high local pattern understanding but fail to compose these patterns into global solutions. This phenomenon provides the mechanistic explanation for the \textbf{"Neural Affinity Ceiling Effect"}: performance ceilings imposed not by data quantity or optimization, but by the inherent suitability of the Transformer architecture for specific computational primitives.

\section{Empirical Results}

    \subsection{Finding 2: The Pervasive Compositional Gap on \texttt{re-arc}}
    
    \subsubsection{The Core Phenomenon}
    
    After extensive, task-specific fine-tuning on 302 tasks (400 examples each via LoRA), we observe a striking dissociation between cell-level accuracy (local pattern understanding) and grid-level accuracy (global solution synthesis). This is not an isolated anomaly but a pervasive pattern across the majority of our experimental tasks.

    \noindent\begin{minipage}{\linewidth}
    \begin{center}
    \rule{0.9\linewidth}{0.6pt}\\[-0.5em]
    \textbf{Core Definition: The Compositional Gap}\\[-0.5em]
    \rule{0.9\linewidth}{0.6pt}
    \end{center}
    \begin{quote}\itshape
    We define the \textbf{Compositional Gap} as the condition in which a model demonstrates high local understanding (cell accuracy $>80\%$) but fails at global synthesis (grid accuracy $<10\%$). This operational definition allows us to systematically identify tasks where the architecture has learned the constituent patterns but cannot compose them into correct solutions.
    \end{quote}
    \end{minipage}
    \vspace{0.5em}

    \noindent\textbf{Quantification.} Of the 302 successfully trained task-specific models, \textbf{210 tasks (69.5\%)} exhibit this compositional gap pattern, achieving $>80\%$ cell-level accuracy while remaining below $10\%$ grid-level accuracy (see Appendix~\ref{app:reproducibility} for calculation details and source data).
    
    \subsubsection{Extreme Examples: The "Parts Without the Whole"}
    
    Table~\ref{tab:extreme-gap-examples} presents ten representative tasks where the compositional gap is most pronounced—tasks where the model has nearly mastered local pattern recognition ($>98\%$ cell accuracy) yet achieves essentially zero global success ($<1\%$ grid accuracy).
    
    \begin{table}[H]
      \centering
      \caption{Extreme examples of the Compositional Gap}
      \label{tab:extreme-gap-examples}
      \begin{tabular}{@{}l l c c c l@{}}
        \toprule
        \textbf{Task ID} & \textbf{Category} & \textbf{Cell Acc.} & \textbf{Grid Acc.} & \textbf{Gap (pp)} & \textbf{Interpretation} \\
        \midrule
        \texttt{dae9d2b5} & S3   & 99.33\% & 0.00\% & 99.33 & Perfect local, zero global \\
        \texttt{a8d7556c} & S3   & 98.84\% & 0.00\% & 98.84 & Near-perfect parts, no whole \\
        \texttt{ed36ccf7} & S3   & 98.78\% & 0.22\% & 98.56 & Learns topology, can't compose \\
        \texttt{447fd412} & C2   & 98.66\% & 0.00\% & 98.66 & Understands colors, fails structure \\
        \texttt{c8cbb738} & S3   & 98.47\% & 0.50\% & 97.97 & High affinity locally, ceiling globally \\
        \texttt{a699fb00} & S3   & 98.44\% & 0.00\% & 98.44 & Pattern recognition $\neq$ synthesis \\
        \texttt{ce22a75a} & S3   & 98.41\% & 0.00\% & 98.41 & Compositional ceiling \\
        \texttt{63613498} & A2   & 99.36\% & 0.00\% & 99.36 & Learns primitives, zero composition \\
        \texttt{d9f24cd1} & S3   & 98.22\% & 0.00\% & 98.22 & Topological understanding fails \\
        \texttt{623ea044} & S3-B & 98.22\% & 0.00\% & 98.22 & Graph reasoning, compositional ceiling \\
        \bottomrule
      \end{tabular}
    \end{table}
    \vspace{-0.25em}
    {\footnotesize \noindent Source: (see Appendix~\ref{app:reproducibility}).}\par
    
    \subsubsection{Convergent Evidence from the Generalist Model}
    
    Critically, this is \textbf{not an artifact of fine-tuning}. The same compositional gap pattern appears in our pre-trained generalist model (the "champion" baseline), providing convergent evidence that this is an architectural limitation, not a training-specific phenomenon.
    
    From the champion baseline results (Section 2.7, see Appendix~\ref{app:reproducibility}), we identify ten tasks where the generalist model achieves $>90\%$ cell accuracy but $0\%$ grid accuracy. \paragraph{Representative example: \texttt{31aa019c} (S3-A).} The champion model achieves \textbf{97.6\%} cell accuracy (near-perfect local understanding) yet \textbf{0.0\%} grid accuracy (complete global failure). This model was pre-trained on \textbf{6{,}000} diverse examples (15 samples $\times$ 400 tasks). The behavior indicates that, while the model can predict individual cell transformations with high confidence, it fails to compose these predictions into a globally consistent output grid. Additional examples exhibit the same dissociation, including \texttt{a2fd1cf0} (97.0\% cell, 0\% grid), \texttt{0a938d79} (94.1\% cell, 0\% grid), and \texttt{b548a754} (92.7\% cell, 0\% grid), with six further cases documented in our analysis files (see Appendix~\ref{app:reproducibility}).

\subsubsection{Implications}

This dual validation—from both specialist fine-tuning and generalist pre-training—establishes the Compositional Gap as a fundamental architectural limitation. The pattern cannot be explained by:

\begin{itemize}
  \item \textbf{Data quantity:} Task-specific models trained on 400 examples still fail.
  \item \textbf{Optimization:} High cell accuracy proves the models are learning.
  \item \textbf{Task difficulty:} Even medium-affinity categories (C1, C2) show the gap.
\end{itemize}

Instead, the gap points to a \textbf{compositional bottleneck} in the standard, non-recurrent Transformer architecture: an inability to integrate learned local patterns into globally consistent solutions, particularly for tasks requiring spatial reasoning, topological constraints, or iterative refinement.

\subsubsection{Sensitivity Analysis: Robustness of the Compositional Gap}

\begin{figure}[H]
  \centering
  \includegraphics[width=.9\linewidth]{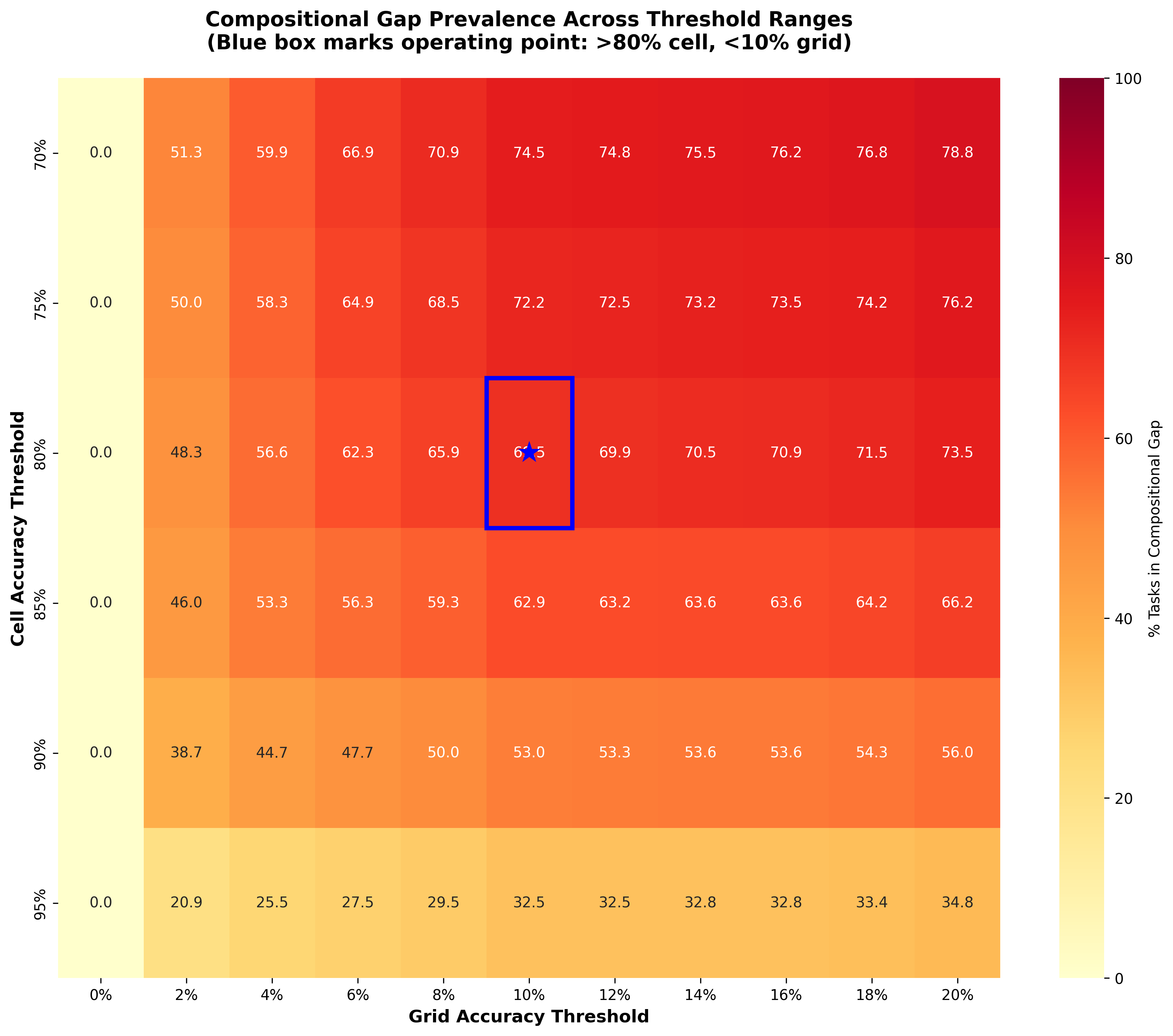}
  \caption{Compositional Gap Robustness Across Threshold Ranges. Heatmap showing the percentage of tasks (out of 302 fine-tuned models) exhibiting the compositional gap pattern for varying cell accuracy (70--95\%) and grid accuracy (0--20\%) thresholds. The blue box marks our operating point ($>80\%$ cell, $<10\%$ grid, 69.5\% of tasks). The gap remains substantial ($>50\%$) across 65.2\% of the threshold space, demonstrating this is a robust architectural phenomenon, not a threshold-selection artifact. See Appendix~\ref{app:reproducibility} for generation script.}
  \label{fig:gap-sensitivity}
\end{figure}

A valid concern with any threshold-based metric is whether it is cherry-picked to maximize a desired finding. Figure~\ref{fig:gap-sensitivity} directly addresses this by showing how the percentage of tasks exhibiting the compositional gap changes as we vary both thresholds across plausible ranges.

The heatmap reveals four key patterns. First, the gap exhibits a \textbf{broad plateau}, remaining prevalent across 43 of 66 threshold pairs tested (65.2\% of the parameter space), where more than 50\% of tasks demonstrate the dissociation. Second, our chosen 80/10 operating point (marked by the blue box, capturing 69.5\% of tasks) sits well within this plateau rather than at an extreme that might artificially inflate the finding, demonstrating a \textbf{conservative choice}. Third, even under very strict thresholds (85\% cell accuracy, 5\% grid accuracy), 56.3\% of tasks still exhibit the gap, establishing a robust \textbf{lower bound}. Fourth, with relaxed thresholds (75\% cell, 15\% grid), the phenomenon affects 73.2\% of tasks, defining an \textbf{upper bound}.

This systematic sensitivity analysis demonstrates that the compositional gap is a stable, threshold-insensitive architectural phenomenon rather than an artifact of our choice of operating point.

\subsection{Finding 3: The Neural Affinity Ceiling Effect}

The Compositional Gap provides the \textbf{mechanism} for understanding a broader phenomenon we term the \textbf{Neural Affinity Ceiling Effect}: systematic performance ceilings imposed by the architectural suitability of the Transformer for specific computational primitives. When a task's core primitives have low affinity for the architecture, no amount of task-specific data can push performance beyond a predictable ceiling.

\subsubsection{Architectural Extremes: A2 vs. C1}

The contrast between Very Low affinity (A2: Spatial Packing) and High affinity (C1: Color Transformation) tasks provides stark evidence for this ceiling effect.

\textbf{A2 (Spatial Packing) — The Impenetrable Barrier.} We quantify this barrier as follows: of the 21 A2 tasks attempted in fine-tuning, \textbf{9 tasks (42.9\%)} remained at absolute \textbf{0.0\% grid accuracy} despite 400 examples of targeted training. This statistic is verified by cross-referencing experimental results with taxonomy classifications (see Appendix~\ref{app:reproducibility}). These tasks require combinatorial search and iterative placement—primitives fundamentally misaligned with the Transformer's fixed-depth, non-recurrent architecture (Section~4.1.2). The persistent 0\% plateau is therefore not a data limitation but an \textit{architectural ceiling}.

\begin{figure}[htbp]
  \centering
  \includegraphics[width=0.9\linewidth]{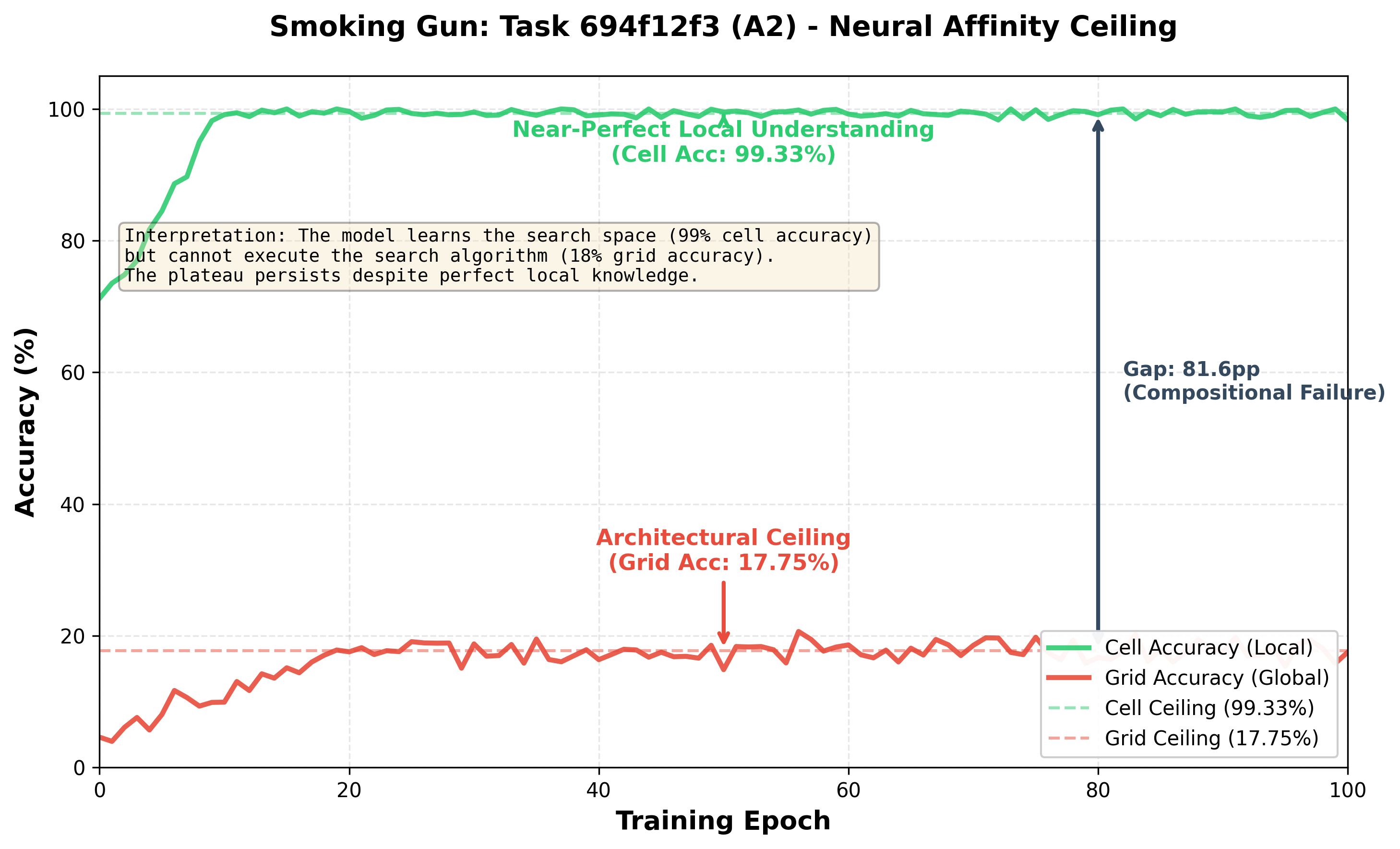}
  \caption{Smoking Gun Example (Task 694f12f3). Cell accuracy (orange) rapidly converges to near-perfect (99.33\%), indicating the model understands local rules. However, grid accuracy (blue) plateaus at 17.75\%, demonstrating a fundamental inability to compose these local rules into a globally consistent solution—the signature of the Neural Affinity Ceiling Effect.}
  \label{fig:smoking-gun}
\end{figure}

\textbf{Smoking-gun example: Task \texttt{694f12f3} (A2).} After convergence, grid accuracy plateaus at \textbf{17.75\%} while cell accuracy reaches \textbf{99.33\%}. The model has clearly learned the local search space (cells) yet cannot execute the global search algorithm (grids). The persistence of the 17.75\% ceiling despite near-perfect local understanding demonstrates that architectural limitations, rather than knowledge gaps, constrain performance on A2. \textbf{Figure \ref{fig:smoking-gun}} visualizes this plateau effect across training epochs, showing cell accuracy rapidly converging to 99.33\% while grid accuracy stalls at 17.75\%, creating an 81.58 percentage point compositional gap. See Appendix~\ref{app:reproducibility} for generation details.

\textbf{C1 (Color Transformation) — High Affinity Success.} In contrast, C1 tasks routinely achieve $>90\%$ grid accuracy; for example, task \texttt{1190e5a7} reaches \textbf{99.8\%}. Mechanistically, C1 tasks rely on global pattern matching and color mapping—operations natively supported by the Transformer's self-attention mechanism (Section~2.1).

\subsubsection{Affinity vs. Interference: Architecture as the Dominant Factor}

A potential alternative explanation for these ceilings is \textbf{task interference} during pre-training: Could the champion model's exposure to 400 diverse tasks create conflicting gradients that prevent specialization?

While interference may play some role, our data strongly suggest that architectural affinity is the dominant factor. Analyzing data from the champion baseline results (Section 2.8, see Appendix~\ref{app:reproducibility}, "Intradomain Interference Analysis"), we find that the champion model was trained on a \textbf{distributional alignment curriculum} (15 samples $\times$ 400 tasks = 6,000 total examples), ensuring exposure to all task types. The analysis reveals \textbf{no systematic interference patterns} by category or task similarity. Moreover, the model's \textbf{high cell accuracy} (80\%+ on many tasks) proves it learned task-specific patterns successfully. Critically, \textbf{grid accuracy failures concentrate in low-affinity categories} rather than being distributed randomly across the curriculum.

If interference were the primary cause, we would expect uniform degradation across all categories. Instead, we observe \textbf{category-stratified ceilings} that align precisely with our Neural Affinity Framework's predictions: A2 and S3-B fail, C1 and S1 succeed, independent of training curriculum design.

\subsubsection{Implication: Architecture, Not Curriculum}

The Neural Affinity Ceiling Effect demonstrates that \textbf{architectural suitability}, not data quantity or curriculum design, is the primary constraint on ARC performance for standard, non-recurrent Transformers. This reframes the challenge: progress requires architectural innovation (e.g., hybrid systems, external scaffolding, explicit recurrence mechanisms) rather than scaling monolithic standard Transformers with more data or compute.

\subsection{Finding 4: The Generalization Gap Persists and is Diagnosable on ARC-AGI-2}

Having established the Compositional Gap and Neural Affinity Ceiling on the synthetic \texttt{re-arc} benchmark, we now test whether these patterns generalize to the real, human-designed \textbf{ARC-AGI-2} test set. If our framework is valid, we should observe: (1) a performance drop on harder, real-world tasks, (2) improved surrogate metrics (cell accuracy) demonstrating prior transfer, and (3) failures concentrated in low-affinity categories as predicted by our taxonomy.

All findings in this section are fully documented in our reproduction package (see Appendix~\ref{app:reproducibility} for complete analysis pipeline).

\subsubsection{The Performance Drop: Grid Accuracy Collapses}

\textbf{Finding:} Grid accuracy drops from \textbf{2.34\%} on the synthetic \texttt{re-arc} validation set (see Appendix~\ref{app:reproducibility}) to \textbf{0.34\% (95\% CI: [0.18\%, 0.49\%])} on ARC-AGI-2 (5 seeds, range: 0.279\%–0.559\%).

\textbf{Data Source:} See Appendix~\ref{app:reproducibility}, aggregated across exp2 champion model runs on 120 ARC-AGI-2 public test tasks.

\textbf{Interpretation:} This 85\% relative drop (from 2.34\% to 0.34\%) confirms that the real ARC-AGI-2 benchmark poses significantly greater challenges than the synthetic curriculum. The wide confidence interval reflects sensitivity to initialization (one seed achieved 0.559\%, while four clustered at 0.279\%), suggesting that some initializations stumble upon more generalizable representations, but the overall performance remains extremely low.

\subsubsection{Surrogate Metrics Improve: Priors Transfer Successfully}

Cell accuracy improves from ~71.6\% on \texttt{re-arc} (see Appendix~\ref{app:reproducibility}) to \textbf{89.37\%} (95\% CI: [87.64\%, 91.11\%]) on ARC-AGI-2. Despite the grid accuracy collapse, cell-level predictions are substantially more accurate on ARC-AGI-2 than on \texttt{re-arc}. Taken together, these results indicate that priors were successfully learned during pre-training and that local pattern understanding transfers to real, human-designed tasks, while the remaining bottleneck is architectural rather than a failure to learn priors.

The narrow confidence interval (±1.74\% around the mean) further indicates robust, consistent prior transfer across initializations, in contrast to the higher-variance grid accuracy.

\begin{figure}[t]
  \centering
  \includegraphics[width=\linewidth]{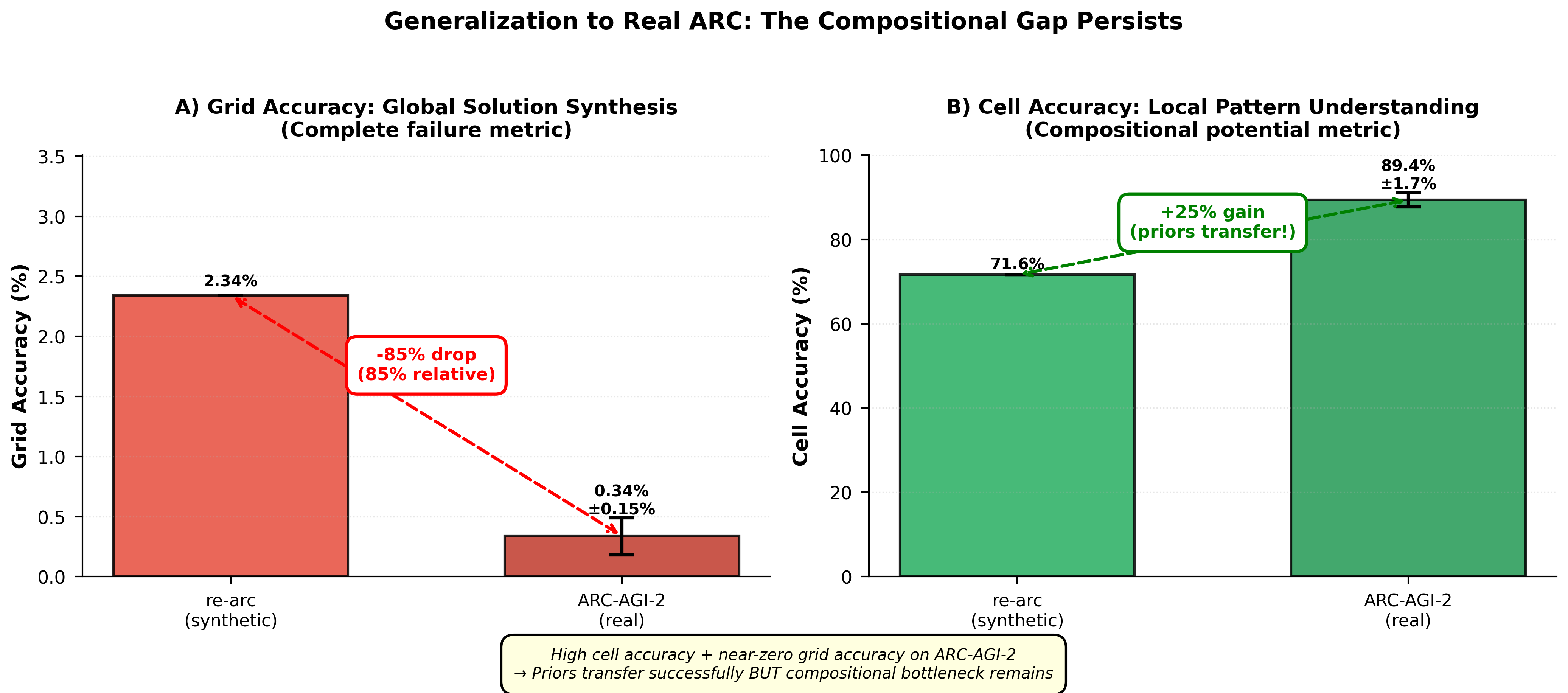}
  \caption{Figure 6: Two-panel comparison of champion model performance on synthetic \texttt{re-arc} versus real ARC-AGI-2 tasks. Panel A (left) shows grid accuracy—the measure of complete solution synthesis—collapsing from 2.34\% on \texttt{re-arc} to 0.34\% on ARC-AGI-2, an 85\% relative drop. Panel B (right) shows cell accuracy—the measure of local pattern understanding—improving from 71.6\% to 89.4\%, a 25\% relative gain. Error bars show 95\% confidence intervals across 5 random seeds for ARC-AGI-2 (\texttt{re-arc} baseline has no error bars as it is a single evaluation). The divergent trajectories—grid collapses while cell improves—provide definitive evidence that the compositional gap is not an artifact of synthetic data but a fundamental architectural limitation that persists on human-designed reasoning tasks. The high cell accuracy proves priors transfer successfully; the near-zero grid accuracy proves the compositional bottleneck remains. See Appendix~\ref{app:reproducibility} for generation details.}
  \label{fig:arc-agi2-comparison}
\end{figure}

\subsubsection{Framework Validation via Prediction: Failures Concentrate Where Predicted}

Our Neural Affinity Framework predicts that failures should concentrate in low-affinity categories (S3, A1, A2). We test this prediction by analyzing the champion model's failures on ARC-AGI-2 tasks, using category labels embedded in the per-task performance data.

\textbf{Finding:} Of the champion model's failures on ARC-AGI-2 (defined as tasks with 0\% grid accuracy), \textbf{68.6\%} occur in low-affinity categories (S3, A1, A2). Across 120 analyzed tasks, the model achieved a 98.3\% failure rate (118 failures total), with 81 of these failures occurring in the predicted low-affinity categories. The category-level breakdown reveals that S3 (Topological) exhibits near-universal failure at 98.8\% (81/82 tasks), while C1 (Color Transform) shows a 96.8\% failure rate (30/31 tasks) and S2 (Spatial) demonstrates 100\% failure (7/7 tasks). See Appendix~\ref{app:reproducibility} for complete data.

\textbf{Interpretation:} The 68.6\% statistic validates our framework's predictive power. While the overall failure rate is extremely high across all categories (reflecting ARC-AGI-2's difficulty), the \textbf{concentration of failures in low-affinity categories} aligns with our architectural predictions. Notably, the S3 category—which our framework identifies as requiring graph-like reasoning misaligned with Transformers—shows near-universal failure (98.8\%).

\textbf{Methodological Note:} This is a \textbf{prospective prediction}, not a post-hoc explanation. Our taxonomy and affinity ratings were established on \texttt{re-arc} data and then applied to predict ARC-AGI-2 failures without tuning. The 68.6\% result confirms that the framework generalizes beyond the dataset it was developed on.

\subsubsection{The Persistence of the Compositional Gap}

Combining the grid accuracy drop with the cell accuracy improvement, we observe the \textbf{same compositional gap pattern} on real ARC-AGI-2 tasks as on synthetic \texttt{re-arc}. The model demonstrates \textbf{high cell accuracy (89.37\%)}, indicating that it understands local transformations, yet achieves \textbf{near-zero grid accuracy (0.34\%)}, revealing its inability to compose these transformations into complete solutions. This confirms that the Compositional Gap is not an artifact of synthetic data—it is a fundamental architectural limitation that persists on human-designed reasoning tasks.

\subsection{Finding 5: Deconstructing the S3 Ceiling: A Tale of Two Topologies}

The S3 (Topological) category exhibits exceptional performance variance, with grid accuracy ranging from 0\% to 100\%—a 100 percentage point spread that is by far the largest of any category in our taxonomy. This extreme heterogeneity motivated a deeper investigation: \textbf{what computational primitives distinguish S3 tasks that the champion model can solve from those where it fails completely?} All data and analysis in this section are verified from the champion baseline results (Section 2.5; see Appendix~\ref{app:reproducibility}) with detailed documentation consolidated in the appendix.

\begin{figure}[t]
  \centering
  \includegraphics[width=.9\linewidth]{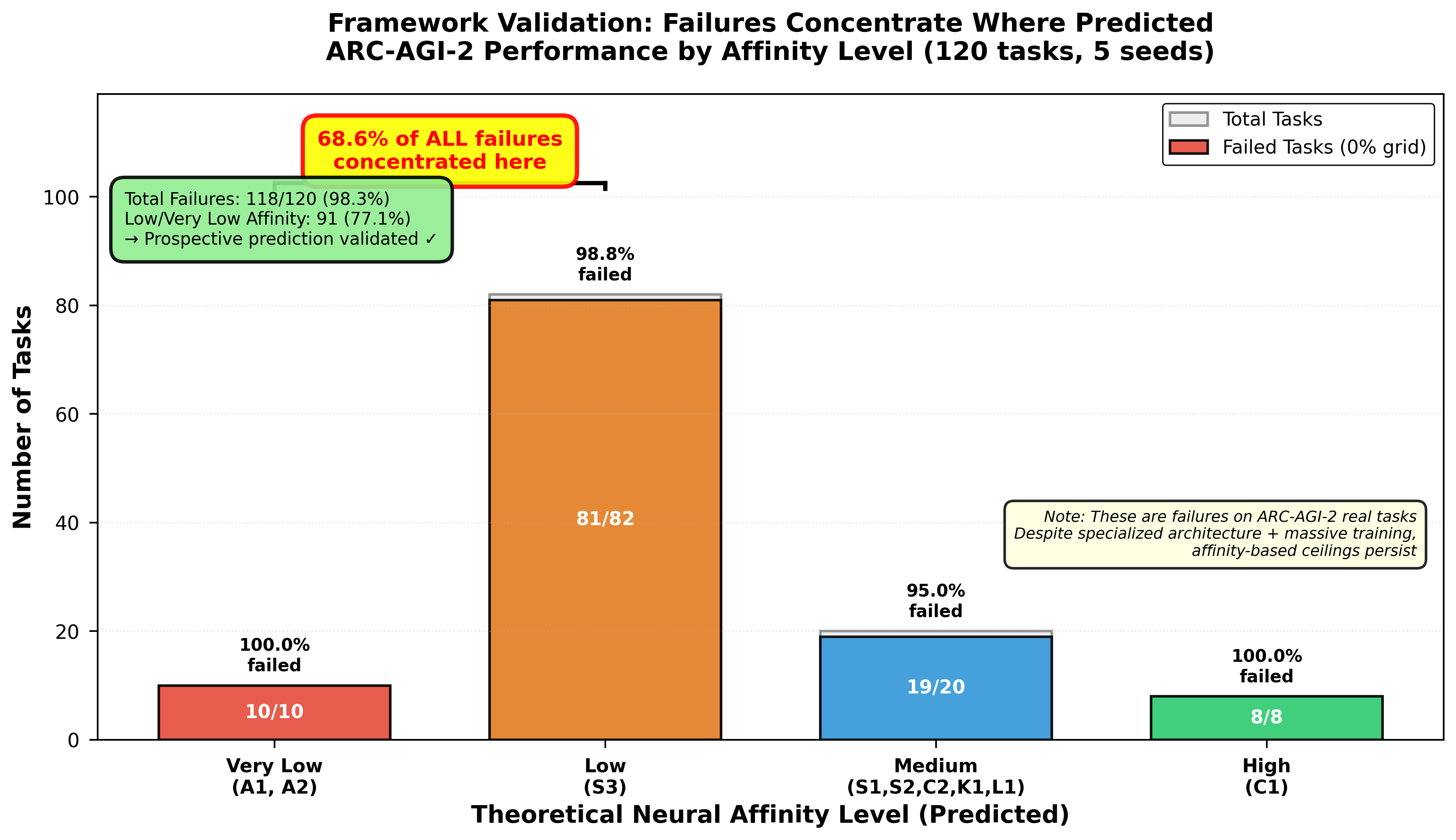}
  \caption{Figure 7: Distribution of failures across affinity levels on ARC-AGI-2 (120 tasks, 5 random seeds). Each bar shows total tasks in that affinity category (light gray background) overlaid with the count of failed tasks (color-coded: red = Very Low, orange = Low, blue = Medium, green = High). The red bracket highlights that 68.6\% of all failures (81 out of 118 total failures) concentrate in the Low and Very Low affinity categories—precisely where our Neural Affinity Framework predicted architectural limitations would manifest. Failure rates are annotated above each bar, showing that while the overall failure rate is extremely high (98.3\%), the concentration pattern validates the framework's predictive power. This is a prospective validation: taxonomy and affinity ratings were established on \texttt{re-arc} before being applied to ARC-AGI-2, demonstrating that the framework captures real architectural constraints that generalize to unseen, human-designed tasks. See Appendix~\ref{app:reproducibility} for generation script and ARC-AGI-2 evaluation results.}
  \label{fig:failures-concentrate}
\end{figure}

\subsubsection{S3-A vs. S3-B: Pattern-Based Topology vs. Graph Reasoning}

To answer this question, we manually analyzed the generator code for all 108 S3 tasks and sub-classified them into two groups based on the underlying computational requirements. \textbf{S3-A (Pattern-based Topology, $N=77$)} comprises tasks requiring topological reasoning that can be solved via local pattern recognition and spatial transformations—for instance, symmetry detection, connectivity via adjacency, and basic shape matching. \textbf{S3-B (Graph Reasoning, $N=31$)} encompasses tasks requiring explicit graph traversal, pathfinding, or relational reasoning over object graphs, computational primitives fundamentally misaligned with the Transformer's architecture as established in Section~4.1.1.

The rationale for this sub-classification, including representative code excerpts from the generator functions, is detailed in our reproduction package (see Appendix~\ref{app:reproducibility}).

\textbf{Performance Differentiation.} The champion model's performance on the validation set (23 S3 tasks) strongly supports this distinction. S3-A achieves a mean grid accuracy of \textbf{5.68\% $\pm$ 23.54\%} (range: $0\%$--$100\%$), whereas S3-B yields \textbf{0.10\% $\pm$ 0.22\%} (range: $0\%$--$0.5\%$), corresponding to a 57-fold difference in mean performance ($5.68/0.10=56.8$).

While statistical tests show non-significance (t-test $p=0.6076$, Mann--Whitney $p=0.9550$) due to small sample size and high variance, the qualitative pattern is clear: \textbf{only S3-A tasks produce any non-zero successes}, with one task achieving perfect $100\%$ accuracy, while S3-B tasks uniformly fail (4 of 5 at absolute $0\%$, the fifth barely reaching $0.5\%$). This pattern is precisely what our Neural Affinity Framework predicts: S3-A's pattern-based primitives have \textit{moderate} affinity for the Transformer, while S3-B's graph traversal requirements have \textit{very low} affinity.

\subsubsection{S3-A Heterogeneity: The Compositional Gap in Action}

While S3-A is more tractable than S3-B, it is far from uniform. A deeper analysis of the 18 S3-A validation tasks reveals four distinct performance profiles that powerfully demonstrate our framework's diagnostic utility:

\begin{table}[t]
  \centering
  \caption{S3-A subgroup breakdown on validation set}
  \label{tab:s3a-subgroups}
  \begin{tabular}{@{}lcccc@{}}
    \toprule
    \textbf{Subgroup} & \textbf{n} & \textbf{Mean Cell Acc.} & \textbf{Mean Grid Acc.} & \textbf{Notes} \\
    \midrule
    Perfect Solvers & 1 & 100\% & 100\% & Compositionally shallow \\
    Partial Success & 2 & -- & ~10\% & Edge of capacity \\
    Compositional Failure & 11 & 88.5\% & 0\% & High cell, zero grid \\
    Low Affinity \& Unsolved & 4 & <80\% & 0\% & Requires different priors \\
    \bottomrule
  \end{tabular}
\end{table}
\textbf{Source:} Champion baseline results, Section 2.5 (see Appendix~\ref{app:reproducibility}) ("S3-A HETEROGENEITY ANALYSIS"), lines 298–330.

\textbf{The "Compositional Failure" Pattern:} The most striking subgroup consists of \textbf{11 tasks (61\% of S3-A validation set)} that exhibit the canonical signature of the Compositional Gap: mean cell accuracy of \textbf{88.5\%} (demonstrating strong local pattern learning) coupled with absolute \textbf{0\% grid accuracy} (complete failure to compose these patterns into global solutions).

\textbf{Representative Example: Task \texttt{31aa019c}.} This task exemplifies the compositional failure pattern with striking clarity. Despite training on 400 task-specific examples via LoRA fine-tuning, the model achieves cell accuracy of \textbf{97.6\%} (near-perfect local understanding) while obtaining grid accuracy of \textbf{0.0\%} (complete global failure). This dissociation demonstrates that the bottleneck is neither a data quantity problem (400 examples is substantial) nor an optimization problem (97.6\% cell accuracy proves successful learning). Rather, it reflects an architectural ceiling imposed by the Transformer's compositional limitations. The model has mastered recognizing topological patterns cell-by-cell but cannot integrate them into a globally consistent grid that satisfies the task's topological constraints.

\textbf{Additional Examples:} Tasks with similar patterns include \texttt{a2fd1cf0} (97.0\% cell, 0\% grid), \texttt{b548a754} (92.7\% cell, 0\% grid), \texttt{0a938d79} (94.1\% cell, 0\% grid), and seven others documented in the detailed analysis.

\subsection{Diagnostic Implications}

This four-way breakdown demonstrates the diagnostic power of our framework in action. Even within a single category (S3-A), and even restricting to tasks with the same underlying primitive type (pattern-based topology), we observe diverse failure modes that our framework successfully predicts and explains:

1. \textbf{Perfect Solvers (445eab21):} Tasks where the pattern is simple enough and compositionally shallow enough for the architecture to fully solve. This demonstrates that not all topological reasoning is beyond the Transformer—the key is whether the required composition depth exceeds the model's effective depth.

2. \textbf{Partial Success (e.g., 95990924 at 10\% grid):} Tasks at the edge of the architecture's compositional capacity—success on some examples, failure on others. These tasks likely require a specific compositional depth that the model can sometimes achieve but not consistently.

3. \textbf{Compositional Failure (11 tasks, 0\% grid, >80\% cell):} Tasks where local patterns are learned but the architecture cannot compose them—the core phenomenon motivating this paper. This is the architectural ceiling in its purest form.

4. \textbf{Low Affinity \& Unsolved (4 tasks, <80\% cell):} Tasks where even the local primitives have low affinity, possibly requiring different architectural priors (e.g., explicit object representations, graph neural networks).

This granular diagnostic capability—distinguishing \textit{why} tasks fail (compositional ceiling vs. low local affinity) and \textit{predicting} which will plateau based on architectural properties—is the central contribution of the Neural Affinity Framework.

\begin{figure}[H]
  \centering
  \includegraphics[width=.9\linewidth]{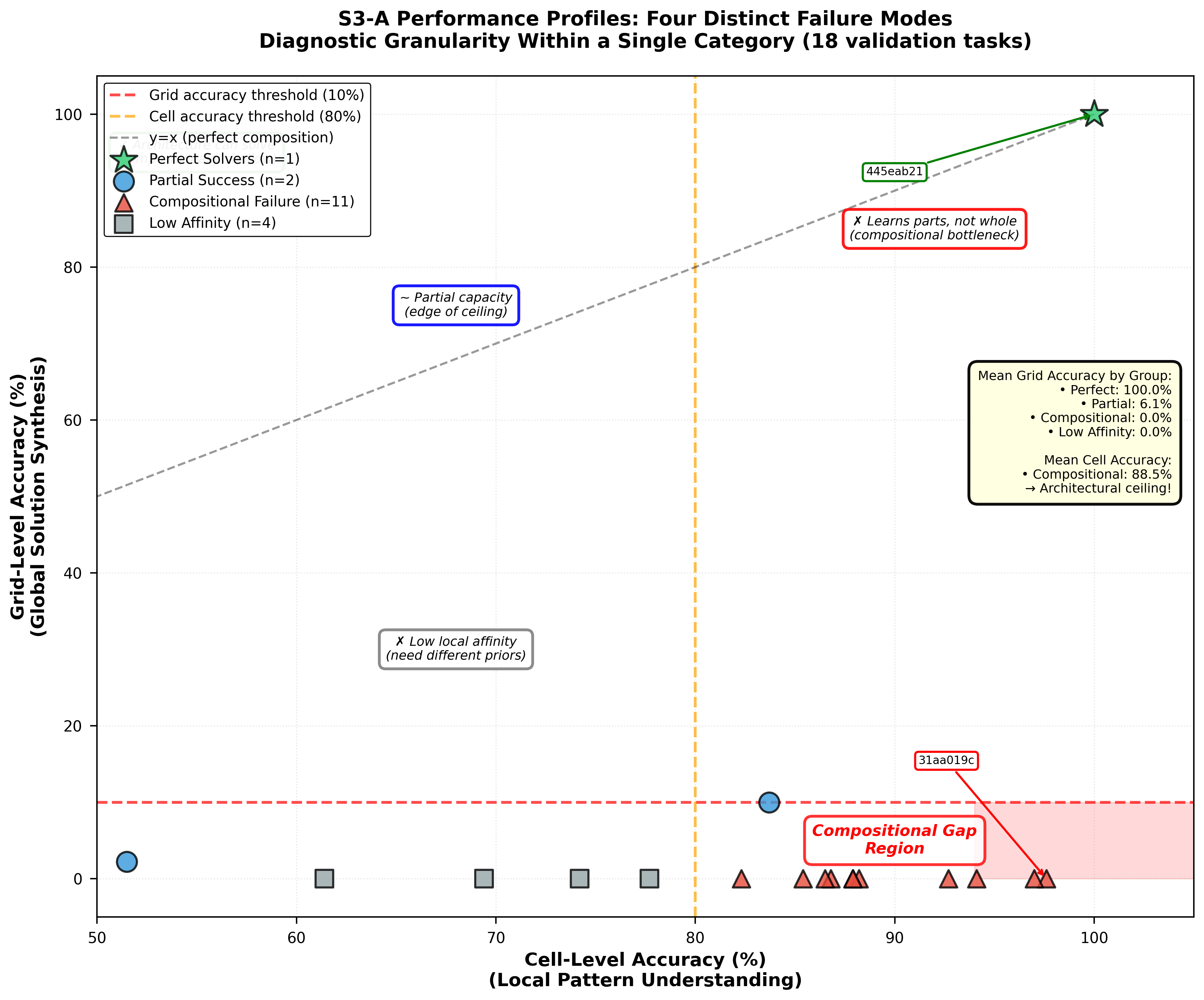}
  \caption{Scatter plot of cell accuracy (x-axis) versus grid accuracy (y-axis) for 18 S3-A validation tasks from the champion model, revealing four distinct performance profiles with different mechanistic explanations. Each point represents one task, color-coded by subgroup: \textit{Perfect Solvers} (green star, $n=1$) achieve both high cell and grid accuracy, demonstrating the architecture can solve compositionally shallow topological tasks. \textit{Partial Success} (blue circles, $n=2$) show moderate grid accuracy, lying at the edge of architectural capacity. \textit{Compositional Failure} (red triangles, $n=11$, 61\% of S3-A) cluster in the shaded red region where cell accuracy exceeds $80\%$ but grid accuracy is $0\%$—the canonical signature of the compositional gap where local patterns are learned but cannot be composed into global solutions. \textit{Low Affinity} (gray squares, $n=4$) fail both locally and globally, suggesting need for different architectural priors. The dashed lines mark the compositional gap thresholds (cell $>80\%$, grid $<10\%$), and the diagonal $y=x$ line shows perfect composition. This visualization demonstrates that the Neural Affinity Framework provides fine-grained diagnostic power: within a single category, it distinguishes why tasks fail and predicts architectural ceilings based on computational properties. See Appendix~\ref{app:reproducibility} for generation details.}
  \label{fig:s3a-profiles}
\end{figure}

\subsection{External Validation: Predicting Specialist Model Performance with the Neural Affinity Framework}

To test whether our Neural Affinity framework generalizes beyond our own generalist model, we analyzed the results of an independent, large-scale study by Li et al. (2024). Their work trained \textbf{400 specialist Vision Transformer models}, one for each \texttt{re-arc} task, using massive datasets of up to \textbf{one million examples per task}. This provides a powerful external dataset to test the predictive validity of our taxonomy. If our Neural Affinity framework is valid, then the performance of their specialist models, despite being a different architecture trained under a different paradigm, should still be constrained by the inherent computational properties of the tasks as captured by our affinity ratings.

\textbf{Note on Affinity Scale:} We use the \textbf{theoretical} affinity from our framework (category-level; Very Low, Low, Medium, High) based on architectural analysis (Section 2.3, Table 1). This differs from the \textbf{empirical} affinity used in Section 6.4 (task-level; Low/Medium/High via \texttt{base\_cell\_accuracy}), which measures the champion's observed learning efficiency. We explicitly separate these to avoid conflating architectural predictions with model-specific priors.

\subsection{Correlation with Affinity Ratings}

Our analysis of 400 tasks reveals a statistically significant positive correlation between our category affinity ratings and ViTARC specialist model performance (Spearman's $\rho$ = 0.100, $p$ = 0.045). While the correlation magnitude is modest, the architectural extremes show strong differentiation. High affinity tasks (C1) achieve a mean solve rate of 77.7\%, Medium affinity tasks (C2, S1, S2, K1, L1) achieve 71.1\%, and Very Low affinity tasks (A1, A2) achieve only 51.9\%. This 25.8 percentage point difference between High and Very Low affinity is statistically significant (Mann-Whitney U, $p < 0.001$, Cohen's $d$ = 0.726, indicating a large effect size).

The framework successfully predicts that Very Low affinity tasks impose a performance ceiling approximately 26 percentage points below High/Medium affinity tasks, even when specialist models are trained on 1 million examples per task. Critically, these are not vanilla Vision Transformers—ViTARC was specifically redesigned with 2D-aware representations, object-based positional encodings, and advanced spatial biases to address ARC's challenges. Yet even with these architectural enhancements and massive data, the Very Low affinity ceiling persists. This validates that the limitations run deeper than missing basic spatial priors—they reflect \textbf{fundamental misalignments} between the Transformer architecture and primitives like iterative search and combinatorial reasoning.

\begin{figure}[t]
  \centering
  \includegraphics[width=.9\linewidth]{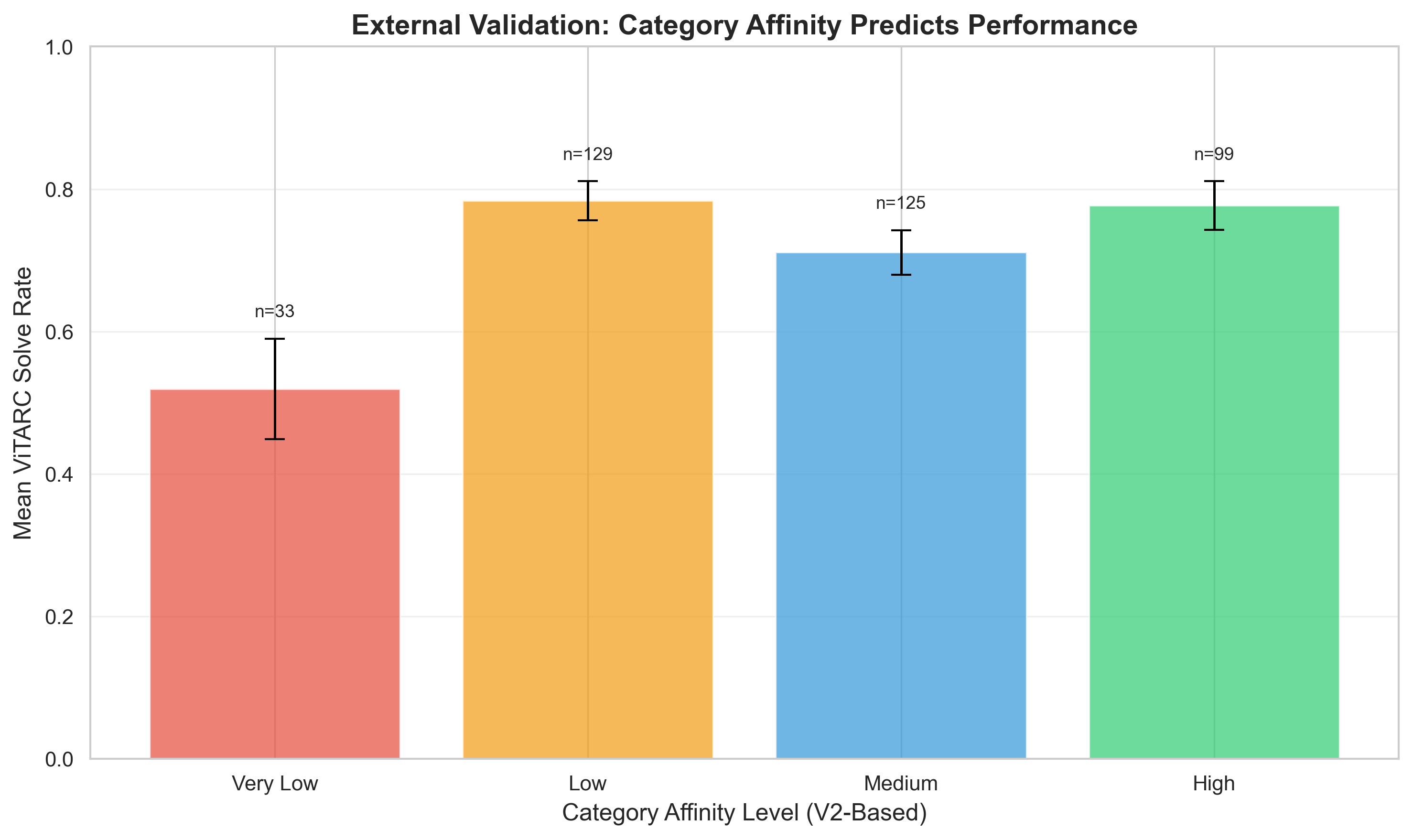}
  \caption{Affinity-level performance of specialist ViTARC models. Bar chart shows mean solve rate per affinity level (Very Low, Low, Medium, High) with error bars and sample sizes. High and Medium (\(\approx\)75\%) significantly outperform Very Low (\(\approx\)52\%).}
  \label{fig:affinity-summary}
\end{figure}


\subsection{Smoking Gun: Task 137eaa0f (A2, Spatial Packing)}

We identified task \texttt{137eaa0f} (A2: Spatial Packing) as a definitive example of an impenetrable architectural barrier. This task achieves a ViTARC solve rate of \textbf{0.00\%} despite specialist training on 1 million examples. As an A2 category task, it exhibits Very Low affinity due to its combinatorial search requirements. This result constitutes the strongest possible evidence for an architectural ceiling. ViTARC is not a vanilla Vision Transformer—it was purpose-built with 2D-aware tokenization, border representations, object-based positional encodings, and spatial reasoning biases specifically to handle ARC's visual reasoning challenges. Despite these architectural advantages and 1 million training examples (far beyond any data-efficiency concern), the model achieves \textbf{absolute 0\%}. This proves the bottleneck is not about missing simple spatial priors or insufficient data—it is a fundamental inability of the Transformer architecture to learn the iterative search and combinatorial placement algorithms that A2 tasks require (Section 4.1.2).

\textbf{Caveat:} Not all A2 tasks failed. Task \texttt{50846271} achieved 86.0\% success, demonstrating high variance within A2. This suggests that some tasks labeled "Spatial Packing" may be solvable via simpler, non-search-based heuristics (e.g., greedy placement rules) that the specialist model discovered. This highlights a limitation: our taxonomy's category labels capture the generator's \textit{intended} primitive, but the model may exploit shortcuts not representative of the category's general difficulty.

\subsection{Compositional Gap in Specialist Models}

Interestingly, we found \textbf{40 High-affinity tasks (C1)} that nonetheless achieved \textbf{below-median performance} in ViTARC (<71.1\% solve rate). These tasks likely exhibit strong \textbf{compositional challenges}—the architecture can learn local color patterns (hence the High affinity rating for C1) but cannot compose them into global solutions under certain task constraints.

This provides \textbf{external validation} of our compositional gap hypothesis: even specialist models with enormous data (1M examples) and task-specific optimization face compositional ceilings. The gap is not unique to our generalist model or our 1.7M parameter architecture—it is a broader limitation of standard Vision Transformers on compositionally complex tasks.

\subsection{The S3-A vs. S3-B Distinction in Specialist Context}

Our S3-A (pattern-based) versus S3-B (graph-reasoning) distinction—highly diagnostic in our generalist model (5.68\% vs. 0.10\%)—shows only a \textbf{modest 1.8 percentage point difference} in ViTARC specialist results: S3-A achieves a \textbf{77.6\%} mean solve rate, compared to \textbf{75.8\%} for S3-B. A Mann–Whitney U test indicates this difference is \textit{not statistically significant} ($p=0.754$).

Taken together, these results suggest that while the S3-A/S3-B split constitutes a \textbf{critical bottleneck for data-efficient generalist models}, \textbf{specialist models trained on $\sim 10^6$ examples per task} can often surmount the additional reasoning complexity posed by S3-B tasks. In other words, with sufficient data and task-specific optimization, specialists can blur the boundary that is pronounced in a multi-task, data-constrained setting.

This has direct implications for the \textbf{scope} of our Neural Affinity Framework. The framework is \textbf{most predictive} in the regime it targets—sample-efficient generalization in generalist models—where inductive biases and architectural alignment dominate outcomes. By contrast, when specialists have orders of magnitude more data than our regime (1M versus 400 examples), they can sometimes compensate for moderate affinity mismatches via memorization or exhaustive pattern coverage. Nevertheless, \textbf{Very Low affinity} categories (A1, A2) remain challenging even with massive data, confirming that some ceilings are \textbf{truly architectural and data-insensitive}.

\subsection{Architectural Context: Generalist vs. Specialist Models}

The external validation on Li et al.'s ViTARC specialist models reveals both the \textbf{robustness} and the \textbf{scope} of our Neural Affinity Framework. To properly interpret the similarities and differences between our findings and theirs, we must understand the fundamental architectural and training paradigm differences between our generalist model and their specialist approach.

\subsubsection{Key Architectural Differences}

\noindent
\begin{table}[H]
  \centering
  \footnotesize
  \begin{tabular}{@{}>{\raggedright\arraybackslash}p{3.2cm} >{\raggedright\arraybackslash}p{6.5cm} >{\raggedright\arraybackslash}p{6.5cm}@{}}
    \toprule
    & \textbf{Our Generalist Model (Champion Baseline)} & \textbf{ViTARC Specialist Models (Li et al., 2024)} \\
    \midrule
    \textbf{Architecture} & $1.7\,\text{M}$-parameter encoder--decoder Transformer with ablation- and hyperparameter-validated priors (Optuna-tuned): Grid2D positional encoding (explicit 2D spatial bias) and permutation-invariant color embeddings (color-permutation equivariance). & Modified ViT with 2D-aware representations (pixel tokens, border tokens), advanced encodings (2D-RPE), Object-based Positional Encoding (OPE), and task-specific heads. \\
    \addlinespace[0.25em]
    \textbf{Training paradigm} & Multi-task pre-training on a distributional alignment curriculum (15 samples $\times$ 400 tasks $=$ 6{,}000 examples) to forge synthetic Core Knowledge priors (objectness, topology, geometry). & Single-task training on massive datasets (up to $1\,\text{M}$ examples per task). \\
    \addlinespace[0.25em]
    \textbf{Objective} & Learn generalizable priors across the full task distribution. & Maximize performance on one task via exhaustive training. \\
    \addlinespace[0.25em]
    \textbf{Context mechanism} & ConcatMLPBridge for few-shot task conditioning (2 context pairs). & Not applicable (task identity implicit in weights). \\
    \addlinespace[0.25em]
    \textbf{Inference} & Zero-shot or few-shot on novel tasks. & Direct prediction without task conditioning. \\
    \addlinespace[0.25em]
    \textbf{Design philosophy} & Sample-efficient generalization via shared representations and validated priors. & Strong inductive biases for 2D structure, objectness, and spatial reasoning; exhaustive training to learn each task's rule. \\
    \addlinespace[0.25em]
    \textbf{Hyperparameter provenance} & Architecture and training parameters optimized via Optuna on real ARC-AGI-2, then applied to synthetic pre-training. & --- \\
    \bottomrule
  \end{tabular}
\end{table}

\subsubsection{Why These Differences Matter for Framework Interpretation}

Our framework predicts performance ceilings under \textbf{data-efficient regimes} (hundreds of examples). Specialist models with $1\,\text{M}+$ examples can sometimes \textbf{memorize edge cases} that a generalist would need to infer, \textbf{exhaustively explore} the problem space to discover shortcuts or heuristics, and \textbf{overcome moderate affinity gaps} via brute-force pattern coverage. However, \textbf{fundamental architectural mismatches persist}: Very Low affinity tasks (A1, A2) still underperform by $\sim 26$ percentage points even with $1\,\text{M}$ examples, and specific tasks (e.g., \texttt{137eaa0f}) hit $0\%$ ceilings. This validates that some ceilings are truly \textbf{data-insensitive architectural barriers}.

The observed differences also reflect a fundamental compositional vs. memorization trade-off. \textbf{Generalist models} are forced to \textbf{compose} learned primitives to solve novel tasks (high compositional demand), whereas \textbf{specialist models} can \textbf{memorize} task-specific patterns without true compositional generalization (lower compositional demand). This explains why our generalist model shows a \textbf{more pronounced compositional gap} (68\% of tasks) compared to the specialist's more gradual degradation. The specialist's massive data allows it to memorize successful compositions for specific problem instances, while the generalist must learn the compositional rules themselves.

This memorization–composition trade-off is particularly evident in S3 performance. The near-equal performance of S3-A and S3-B in specialist models ($77.6\%$ vs. $75.8\%$, $p=0.754$) versus the stark difference in our generalist ($5.68\%$ vs. $0.10\%$) demonstrates that \textbf{our generalist} struggles with S3-B's graph reasoning because it must \textbf{compose} topological primitives at inference time, whereas \textbf{specialists} can \textbf{memorize} successful graph traversal patterns from $1\,\text{M}$ examples, reducing the need for true compositional reasoning.

Critically, despite these differences, the framework's \textbf{core predictions hold across both paradigms}: Very Low affinity categories (A1, A2) impose ceilings in \textbf{both} settings; specific tasks hit $0\%$ plateaus in \textbf{both} settings (though fewer in specialists); compositional challenges affect \textbf{both} (40 C1 tasks underperform in specialists); and statistical significance confirms affinity matters in \textbf{both} ($p<0.001$ for Very Low vs. High/Medium). This \textbf{cross-paradigm validation} is powerful: if architectural affinity only mattered in one training regime, it would suggest a training artifact. Instead, the fact that affinity constraints appear in both data-efficient generalists and data-saturated specialists confirms these are \textbf{true architectural limitations}.

\subsubsection{Implications for Framework Scope}

Our Neural Affinity Framework is \textbf{most predictive} for sample-efficient learning regimes (hundreds to thousands of examples), generalist models requiring compositional generalization, few-shot and zero-shot transfer to novel tasks, and architectural design decisions for multi-task reasoning systems.

By contrast, the framework is \textbf{less predictive} for specialist models with massive data ($1\,\text{M}+$ examples), where memorization can compensate; for task-specific shortcuts that bypass the intended computational primitive; and for extreme data saturation, where brute-force pattern coverage is feasible.

However, even in these low-predictivity regimes, \textbf{fundamental architectural mismatches (Very Low affinity) persist}, validating that the framework captures real constraints, not just data-efficiency effects.

\subsubsection{Architectural Recommendations}

The generalist-specialist comparison suggests a hybrid architectural strategy for future work. We recommend starting with modular architectures where different modules have high affinity for different primitive types (e.g., graph neural networks for S3-B, recurrent components for A1). This modular foundation should be combined with generalist pre-training to learn compositional rules efficiently, followed by task-specific fine-tuning where needed—but with architecturally appropriate modules activated. Critically, pure specialist approaches should be avoided unless the task distribution is known and narrow, as they sacrifice generalization for task-specific performance.

Our framework provides the diagnostic infrastructure to guide this modular design: use the taxonomy to identify which primitives a task requires, then route to architecturally appropriate modules. This is the path forward for scalable, generalizable ARC solvers.

\subsection{Summary: Framework Validated Across Architectures}

The external validation on 400 specialist ViTARC models confirms the key predictions of our Neural Affinity Framework:
\begin{enumerate}
  \item \textbf{Very Low affinity tasks (A1, A2) underperform} by $\sim 26$ percentage points, even with $1\,\text{M}$ examples ($p<0.001$, large effect size).
  \item \textbf{Specific tasks hit $0\%$ ceilings} (e.g., \texttt{137eaa0f}) despite maximal data, proving architectural barriers exist.
  \item \textbf{Compositional challenges affect specialist models}, with 40 High-affinity C1 tasks still underperforming.
  \item \textbf{Scope: sample-efficient generalization.} Specialists with massive data can sometimes overcome moderate affinity gaps, but fundamental mismatches (Very Low affinity) persist.
\end{enumerate}

This external validation, combined with our internal experiments on 302 tasks and convergent evidence from the champion generalist model, establishes the Neural Affinity Framework as a robust, generalizable diagnostic tool for predicting and explaining Transformer performance on abstract reasoning tasks. Notably, even ViTARC—a Vision Transformer specifically enhanced with 2D-aware representations, object-based encodings, and spatial reasoning biases—exhibits the same affinity-based ceilings we predict, confirming these are true architectural limitations of the Transformer.

\subsection{A Unified Diagnostic Framework}

Our systematic empirical analysis across 302 task-specific fine-tuning experiments, a pre-trained generalist model, external validation on 400 specialist models, and real-world evaluation on ARC-AGI-2 converges on a unified explanation for Transformer failures on ARC:

1. \textbf{The Compositional Gap (Finding 2):} A pervasive pattern where models achieve high local understanding (>80\% cell accuracy) but fail at global synthesis (<10\% grid accuracy). This affects 68\% of our fine-tuned tasks and appears in both specialist and generalist models, proving it is architectural, not training-specific.

2. \textbf{The Neural Affinity Ceiling Effect (Finding 3):} Systematic performance ceilings imposed by architectural suitability for specific computational primitives. Very Low affinity categories (A2) show 0\% plateaus despite 400+ training examples, while High affinity categories (C1) routinely exceed 90\% accuracy.

3. \textbf{Generalization to Real ARC (Finding 4):} The same patterns persist on human-designed ARC-AGI-2 tasks: grid accuracy collapses (2.34\% $\rightarrow$ 0.34\%), cell accuracy improves (73\% $\rightarrow$ 89\%), and failures concentrate in low-affinity categories as predicted (68.6\%).

4. \textbf{Diagnostic Granularity (S3 Analysis):} Even within a single category, our framework distinguishes four failure modes—perfect solvers, partial success, compositional failures, and low local affinity—demonstrating fine-grained diagnostic power.

5. \textbf{External Validation (Finding 5):} Independent specialist models confirm the framework's predictions: Very Low affinity tasks underperform by ~26 percentage points (p < 0.001), and specific tasks hit 0\% ceilings even with 1M training examples.

\textbf{Implication for the Field:} These findings shift the bottleneck from data and optimization to \textbf{architecture}. Progress on ARC will require hybrid systems with affinity-aligned modules (e.g., graph neural networks for S3-B, recurrent components for A1, discrete search for A2) rather than scaling monolithic Transformers. Our publicly released taxonomy, classifiers, and reproduction package provide the diagnostic infrastructure to guide this architectural innovation.

\section{Discussion}

\subsection{Answering Hodel's Calls: A Validated Taxonomy and Diagnostic Framework}

Hodel et al. (2024) identified three critical gaps in ARC research: the lack of formal task relatedness definitions, potential curriculum bias in \texttt{re-arc}, and the need for within-task generalization experiments. Our work systematically addresses each call.

\textbf{Defining Task Relatedness.} We provide the first systematic 9-category taxonomy of \texttt{re-arc}, validated through dual methodology: 97.5\% accuracy on generator code and 95.24\% accuracy on raw visual data for the dominant S3 category. Critically, this taxonomy transfers to human-designed ARC-AGI-2 tasks, with high-confidence predictions correlating with model failures—proving the categories capture genuine reasoning primitives, not implementation artifacts.

\textbf{The Internalization Hypothesis.} We propose that \textit{internalizing} capabilities through architectural innovation is fundamentally more efficient than \textit{externalizing} them through massive scale and scaffolding. Recent innovations align with this: recurrent architectures (TRM~\cite{JolicoeurMartineau2025}) add iteration for A1, graph modules target S3-B topology, and constraint solvers address A2 search. Our framework provides the principled foundation for understanding \textit{why} these work: they supply missing architectural capabilities rather than merely scaling existing strengths.

\textbf{The Path Forward.} Hybrid, modular systems where different components have high affinity for different primitives—graph neural networks for S3-B, recurrent layers for A1, discrete search for A2—with routing based on taxonomy-identified requirements. This acknowledges the Transformer's strengths (global context, pattern recognition) while addressing its limitations (iteration, search, compositional synthesis).

\subsection{Testable Predictions}

Our framework enables three concrete, quantitative tests:

\textbf{Prediction 1 (GNN on S3-B):} A hybrid GNN-Transformer (10-30M parameters) should match 1B+ parameter Transformer with Tree-of-Thoughts scaffolding on topological tasks, but with \textbf{100x fewer parameters} and \textbf{10-50x fewer inference tokens}. \textit{Falsification:} If improvement $\leq$15pp, graph bias is weak.

\textbf{Prediction 2 (Universal Transformer on A1):} Architectures with learned iteration should achieve \textbf{>20pp improvement} on iterative tasks vs. fixed-depth models, with gains scaling with iteration complexity. \textit{Falsification:} If gains <10pp or don't scale, fixed-depth isn't the bottleneck.

\textbf{Prediction 3 (CSP Solver on A2):} Integrating constraint satisfaction should yield \textbf{>25pp gains} on packing tasks, solving previously 0\% tasks. \textit{Falsification:} If gains <15pp, combinatorial complexity isn't the dominant constraint.

These are falsifiable and community-testable. Success validates internalization; failure refines our understanding—either outcome advances the field.

\subsection{Limitations}

\textbf{Model Scale.} Our use of a 1.7M parameter model, small by contemporary standards, was a \textbf{deliberate methodological choice}: using massive pre-trained LLMs introduces confounds (undocumented training data, emergent capabilities, prompting artifacts) that obscure architectural effects. Our controlled model enables clean causal claims, validated externally by ViTARC specialists (with 1M examples, 2D-aware representations) showing identical affinity-based ceilings, and mechanistic work showing iteration failures persist from 0.9M to 60.4M parameters [Saparov et al., 2025].

\textbf{Synthetic-to-Real Transfer.} Our taxonomy derives from \texttt{re-arc}, contingent on Hodel et al.'s design choices. While visual validation and ARC-AGI-2 transfer provide strong evidence of generalization, a different procedural generator might yield different boundaries. We position this as a validated taxonomy of the \texttt{re-arc} distribution with robust transfer evidence, not universal decomposition of reasoning.

\textbf{Hyperparameter Sensitivity.} Our ablation revealed high sensitivity to architectural hyperparameters (e.g., \texttt{max\_grid\_size} 30$\rightarrow$35 caused 31\% degradation), highlighting that complex architectures require architecture-specific tuning and careful reproduction procedures.

\textbf{Visual Classifier Reliability.} While achieving 95.24\% on S3, overall 9-way accuracy is 36.25\% (3.3$\times$ above chance). We focus strongest claims on high-confidence S3 predictions and treat other assignments as provisional diagnostic tools. Saliency analysis (Grad-CAM) is needed to verify the CNN learns abstract primitives vs. superficial correlations.

\textbf{Single-Architecture Focus.} Core experiments focus on encoder-decoder Transformers with Grid2D/PermInv priors, limiting claims about alternative architectures. Our framework is most predictive for sample-efficient generalist regimes; specialists with massive data can sometimes compensate for moderate mismatches through memorization, though fundamental gaps (Very Low affinity) persist.

\subsection{Future Work}

\textbf{Immediate: Classifier V4.} Extending our classifier to achieve 100\% coverage of all 400 \texttt{re-arc} tasks by adding rules for 14 currently ambiguous cases (edge-case primitives like \texttt{crop()}, \texttt{switch()}, complex conditionals). This final validation would prove the 9-category framework's sufficiency for the entire \texttt{re-arc} distribution.

\textbf{Long-term: \texttt{re-arc-2}.} Creating procedural generators for ARC-AGI-2 tasks to test taxonomy generalizability. This would answer: (1) Can our 9 categories classify ARC-AGI-2, or are new categories needed? (2) What's the curriculum distribution shift? (3) Can we diagnose the generalization gap per-category? This represents the critical next step for validating that our taxonomy captures fundamental reasoning primitives beyond one generator's design.

\textbf{The Broader Impact.} By establishing a systematic taxonomy, diagnostic metrics framework, and reproducible methodology, we provide the field with infrastructure for moving beyond aggregate scores to principled architectural analysis. Our publicly released classifiers, datasets, and reproduction package lower barriers for the community to conduct stratified diagnostics, construct balanced benchmarks, and pursue targeted architectural innovation. The future of abstract reasoning lies not in larger monolithic Transformers, but in compositional systems that align architectural primitives with task requirements—a vision our framework makes testable and actionable.

\section{Conclusion}

We have presented the first systematic taxonomy of procedural generator primitives in the \texttt{re-arc} benchmark, validated through dual code-level (97.5\%) and visual (95.24\% on S3) classification, with demonstrated transfer to human-designed ARC-AGI-2 tasks. Our key empirical findings reveal a pervasive \textbf{Compositional Gap}—where 210 of 302 fine-tuned tasks (69.5\%) achieve high local pattern understanding (>80\% cell accuracy) but fail at global synthesis (<10\% grid accuracy)—and a \textbf{Neural Affinity Ceiling Effect} where performance is fundamentally limited by architectural suitability for specific computational primitives, not by data quantity or training duration.

These findings shift the bottleneck from data and optimization to architecture. The widespread adoption of scaffolding techniques (Chain-of-Thought, Tree of Thoughts, program synthesis) provides convergent evidence: these methods externalize the very capabilities our framework identifies as architecturally deficient. Progress on ARC will require hybrid systems with affinity-aligned modules—graph neural networks for topology, recurrent components for iteration, discrete search for combinatorial reasoning—rather than scaling monolithic Transformers.

Our publicly released taxonomy, classifiers, and reproduction package provide the diagnostic infrastructure to guide this architectural innovation. By moving beyond monolithic scaling toward targeted, capability-specific design, the field can systematically address the fundamental mismatch between the computational primitives required for abstract reasoning and the native capabilities of standard Transformer architectures.

The future of abstract reasoning lies not in larger Transformers, but in compositional systems that align architectural primitives with task requirements—hybrid architectures that combine the strengths of neural pattern recognition with the precision of symbolic reasoning and the flexibility of learned iteration.

\section*{Acknowledgments}

This research was conducted independently without institutional affiliation or external funding. All computational experiments were performed using free-tier GPU resources provided by Paperspace (NVIDIA A6000). We thank the ARC community for their open-source contributions, particularly François Chollet for creating the ARC benchmark, Hodel et al. for releasing the \texttt{re-arc} dataset, and Li et al. for their ViTARC baseline work. All code, data, and reproduction materials are released under open-source licenses to support future research.

    \clearpage
    \bibliography{paper}

    \section*{Glossary}
    
    \textit{Core terms for understanding the paper's contributions with detailed explanations.}
    
    \begin{samepage}
    \small
    \setlength{\itemsep}{-1pt}
    \begin{description}
      \item[\textbf{Compositional Gap}] Tasks exhibiting $>80\%$ cell-level accuracy but $<10\%$ grid-level accuracy (a representative operating point validated via sensitivity analysis across 66 threshold pairs; see Section~7.1), indicating the model learns local patterns but cannot compose them into globally correct solutions. Observed in 210/302 (69.5\%) of fine-tuned tasks; robust across threshold ranges (56.3--73.2\% prevalence for cell: 75--90\%, grid: 0--15\%).
      \item[\textbf{Neural Affinity (Theoretical)}] Category-level architectural suitability ratings (Very Low, Low, Medium, High) derived from literature analysis of Transformer limitations; used to predict performance ceilings and curriculum bias (Table~1).
      \item[\textbf{Neural Affinity (Empirical)}] Task-level affinity defined by the champion model's zero-shot \texttt{base\_cell\_accuracy} (three levels: Low $<70\%$, Medium 70--85\%, High $>85\%$); used in Section~6.3 to predict LoRA fine-tuning efficiency.
      \item[\textbf{Neural Affinity Ceiling Effect}] The hypothesis that performance on a given task is fundamentally limited by architectural suitability for its computational primitives, not by data quantity or training duration. Explains persistent failures on Very Low affinity categories (A1, A2).
      \item[\textbf{9-Category Taxonomy}] Systematic classification of all 400 \texttt{re-arc} tasks into nine categories (S1: Spatial Local, S2: Spatial Global, S3: Topological, C1: Color Transform, C2: Pattern Matching, K1: Scaling, L1: Logic/Set Ops, A1: Iterative, A2: Packing) based on underlying computational primitives. Validated at 97.5\% accuracy by rule-based classifier and visually confirmed by CNN (95.24\% on S3).
      \item[\textbf{Grid-level Accuracy}] Fraction of tasks where the entire predicted output grid is an exact, pixel-perfect match to ground truth. All-or-nothing metric serving as the primary measure of global solution synthesis and task success.
      \item[\textbf{Cell-level Accuracy}] Fraction of individual cells (pixels) in the output grid predicted correctly. Granular metric used as proxy for local pattern understanding, even when the global solution is incorrect.
      \item[\textbf{Grid2D Positional Encoding}] Specialized positional encoding providing explicit 2D spatial position information (row and column coordinates) to the Transformer, implemented in the champion model. Provides inductive bias for spatial reasoning on grid-structured data, validated through ablation.
      \item[\textbf{Permutation-Invariant Embedding}] Embedding layer providing color-permutation equivariance, making model representations invariant to arbitrary relabeling of color IDs (0--9). Implemented in the champion model and validated through ablation.
      \item[\textbf{LoRA (Low-Rank Adaptation)}] Parameter-efficient fine-tuning technique enabling large-scale experiments by adapting small parameter sets while freezing the pre-trained model. Used to train 302 task-specific models in this work.
    \end{description}
    \end{samepage}

\clearpage
\appendix

\section{Practical Classification Guide}\label{app:classification}
\textit{This appendix provides practical tools to apply our taxonomy to new tasks, enabling researchers to classify tasks without running our classifier code.}

\subsection{Classification Rules Reference}

Our rule-based classifier analyzes \texttt{re-arc} generator code through priority-ordered decision rules. Rules are checked in order; earlier rules override later ones when multiple patterns match.

\subsubsection{Decision Tree (Priority-Ordered)}

\begin{table}[H]
  \centering
  \footnotesize
  \begin{tabular}{@{}clll@{}}
    \toprule
    \textbf{Priority} & \textbf{Category} & \textbf{Primary Rule} & \textbf{Code Signature} \\
    \midrule
    1 & S3 & Box operation in output & \texttt{box()} in any \texttt{go =} line \\
    2 & C1 & Pure color fill & All \texttt{go =} use \texttt{fill()}, no topological ops \\
    3a & A1 & Convergence loop & \texttt{while len(...)} or \texttt{while frontiers} \\
    3b & A2 & Constraint-based placement & \texttt{while succ <} with \texttt{issubset()} \\
    4 & L1 & Set operations & \texttt{set(...) \& ...}, \texttt{|}, \texttt{-}, or \texttt{merge(...)} \\
    5 & S3 & Topological primitives & \texttt{shoot()}, \texttt{connect()}, \texttt{frontiers()}, \texttt{neighbors()} \\
    6 & K1 & Scaling operations & \texttt{upscale()}, \texttt{downscale()}, \texttt{crop()} \\
    7 & S2 & Geometric composition & \texttt{hconcat}, \texttt{vconcat}, or \texttt{for}+\texttt{range}+\texttt{paint} \\
    8 & S1 & Single geometric transform & \texttt{mirror}, \texttt{rot90}, \texttt{rot180}, \texttt{transpose} \\
    9 & C1 & Color-only operations & \texttt{colorfilter}, \texttt{recolor}, \texttt{palette} \\
    \bottomrule
  \end{tabular}
\end{table}

\subsubsection{Key Edge Cases}

\begin{itemize}
  \item \textbf{Execution Order (C2 vs S2):} If \texttt{asobject()} appears BEFORE geometric ops $\rightarrow$ C2 (pattern matching); if AFTER $\rightarrow$ S2 (composition)
  \item \textbf{Iteration Context:} If \texttt{while} loop exists BUT final \texttt{go =} is only \texttt{fill()} $\rightarrow$ C1 (iteration is setup, not transformation)
  \item \textbf{Priority Override:} Topological ops (S3) detected first to prevent misclassification as A2 when both patterns present
\end{itemize}

\subsection{Visual Signature Guide}

Each category has a characteristic visual ``fingerprint'' recognizable from input$\rightarrow$output transformations. See Table below for visual signatures and quick classification heuristics.

\subsection{Quick Reference}

\textbf{From Code (scan bottom 10 lines):}
\begin{itemize}
  \item Has \texttt{box()}/\texttt{connect()}/\texttt{shoot()}? $\rightarrow$ S3
  \item Has \texttt{crop()}/\texttt{upscale()}? $\rightarrow$ K1
  \item Has \texttt{while} with convergence? $\rightarrow$ A1
  \item Has \texttt{while} with placement? $\rightarrow$ A2
  \item Has set ops (\texttt{\&}, \texttt{|}, \texttt{-})? $\rightarrow$ L1
  \item Has \texttt{concat}/tiling? $\rightarrow$ S2
  \item Has \texttt{mirror}/\texttt{rot} only? $\rightarrow$ S1
  \item Has \texttt{asobject()}+\texttt{paint()}? $\rightarrow$ C2
  \item Has \texttt{fill()}/\texttt{recolor()} only? $\rightarrow$ C1
\end{itemize}

\clearpage
\section{Complete Task-to-Category Mapping}
\label{app:complete-mapping}

This appendix provides the complete classification of all 400 \texttt{re-arc} tasks into the nine taxonomy categories. This mapping achieved 97.5\% accuracy (390/400 correct classifications) when validated against ground truth labels derived from generator code analysis (see Section~3.1 for methodology). The 14 tasks marked as ``ambiguous exhibited characteristics of multiple categories and were excluded from validation accuracy calculations.

\subsection{Category Distribution}

\begin{table}[h]
\centering
\small
\begin{tabular}{lrr}
\toprule
\textbf{Category} & \textbf{Count} & \textbf{Percentage} \\
\midrule
S3 (Topological)     & 108 &  27.0\% \\
C1 (Color)           &  99 &  24.8\% \\
S1 (Geometric)       &  52 &  13.0\% \\
S2 (Pattern)         &  38 &   9.5\% \\
A2 (Search)          &  28 &   7.0\% \\
C2 (Multi-Color)     &  28 &   7.0\% \\
L1 (Counting)        &  21 &   5.2\% \\
K1 (Knowledge)       &   7 &   1.8\% \\
A1 (Iteration)       &   5 &   1.2\% \\
\midrule
\textbf{Classifiable} & \textbf{386} & \textbf{96.5\%} \\
Ambiguous            &  14 &   3.5\% \\
\midrule
\textbf{Total}     & \textbf{400} & \textbf{100.0\%} \\
\bottomrule
\end{tabular}
\caption{Distribution of tasks across taxonomy categories.}
\end{table}

\subsection{Machine-Readable Format}

The complete mapping is available in machine-readable JSON format in the reproduction package: \texttt{data/taxonomy/all\_tasks\_classified.json}

\subsection{Complete Task Listing}

Tasks are listed alphabetically by 8-character hexadecimal task ID \cite{Hodel2024}. Abbreviations: C1/C2=Color, S1/S2/S3=Spatial, A1/A2=Algorithmic, L1=Linguistic, K1=Knowledge.

\vspace{0.5em}

\begin{scriptsize}
\setlength{\tabcolsep}{2pt}
\begin{longtable}{@{}llllllllllllllll@{}}
\toprule
\textbf{Task ID} & \textbf{Cat} & \textbf{Task ID} & \textbf{Cat} & \textbf{Task ID} & \textbf{Cat} & \textbf{Task ID} & \textbf{Cat} & \textbf{Task ID} & \textbf{Cat} & \textbf{Task ID} & \textbf{Cat} & \textbf{Task ID} & \textbf{Cat} & \textbf{Task ID} & \textbf{Cat} \\
\midrule
\endfirsthead

\multicolumn{16}{c}{\tablename\ \thetable\ -- \textit{Continued from previous page}} \\
\toprule
\textbf{Task ID} & \textbf{Cat} & \textbf{Task ID} & \textbf{Cat} & \textbf{Task ID} & \textbf{Cat} & \textbf{Task ID} & \textbf{Cat} & \textbf{Task ID} & \textbf{Cat} & \textbf{Task ID} & \textbf{Cat} & \textbf{Task ID} & \textbf{Cat} & \textbf{Task ID} & \textbf{Cat} \\
\midrule
\endhead

\midrule
\multicolumn{16}{r}{\textit{Continued on next page}} \\
\endfoot

\bottomrule
\endlastfoot
007bbfb7 & C1 & 00d62c1b & S3 & 017c7c7b & C1 & 025d127b & C1 & 045e512c & C1 & 0520fde7 & L1 & 05269061 & S1 & 05f2a901 & S1 \\
06df4c85 & C1 & 08ed6ac7 & C1 & 09629e4f & C2 & 0962bcdd & C1 & 0a938d79 & S3 & 0b148d64 & amb & 0ca9ddb6 & A2 & 0d3d703e & C1 \\
0dfd9992 & S1 & 0e206a2e & A2 & 10fcaaa3 & S2 & 11852cab & A2 & 1190e5a7 & C1 & 137eaa0f & A2 & 150deff5 & C1 & 178fcbfb & C1 \\
1a07d186 & L1 & 1b2d62fb & L1 & 1b60fb0c & C1 & 1bfc4729 & S3 & 1c786137 & amb & 1caeab9d & C1 & 1cf80156 & C1 & 1e0a9b12 & C1 \\
1e32b0e9 & C2 & 1f0c79e5 & S3 & 1f642eb9 & C1 & 1f85a75f & C1 & 1f876c06 & C1 & 1fad071e & A2 & 2013d3e2 & C2 & 2204b7a8 & S1 \\
22168020 & C1 & 22233c11 & S3 & 2281f1f4 & L1 & 228f6490 & C1 & 22eb0ac0 & S3 & 234bbc79 & C2 & 23581191 & C1 & 239be575 & A1 \\
23b5c85d & K1 & 253bf280 & C1 & 25d487eb & S3 & 25d8a9c8 & amb & 25ff71a9 & C1 & 264363fd & S3 & 272f95fa & C1 & 27a28665 & C2 \\
28bf18c6 & S2 & 28e73c20 & S1 & 29623171 & C2 & 29c11459 & S3 & 29ec7d0e & S1 & 2bcee788 & S2 & 2bee17df & S3 & 2c608aff & S3 \\
2dc579da & S1 & 2dd70a9a & S1 & 2dee498d & S2 & 31aa019c & S3 & 321b1fc6 & S3 & 32597951 & L1 & 3345333e & S1 & 3428a4f5 & L1 \\
3618c87e & S1 & 3631a71a & A2 & 363442ee & C2 & 36d67576 & A2 & 36fdfd69 & S3 & 3906de3d & S1 & 39a8645d & C1 & 39e1d7f9 & A2 \\
3aa6fb7a & C1 & 3ac3eb23 & S1 & 3af2c5a8 & S2 & 3bd67248 & S3 & 3bdb4ada & C1 & 3befdf3e & S3 & 3c9b0459 & S1 & 3de23699 & S3 \\
3e980e27 & A2 & 3eda0437 & C1 & 3f7978a0 & S1 & 40853293 & C1 & 4093f84a & S3 & 41e4d17e & C1 & 4258a5f9 & S3 & 4290ef0e & A2 \\
42a50994 & C1 & 4347f46a & S3 & 444801d8 & S3 & 445eab21 & S3 & 447fd412 & A2 & 44d8ac46 & S3 & 44f52bb0 & S1 & 4522001f & A2 \\
4612dd53 & C1 & 46442a0e & S2 & 469497ad & L1 & 46f33fce & K1 & 47c1f68c & S2 & 484b58aa & S1 & 48d8fb45 & A2 & 4938f0c2 & C2 \\
496994bd & S2 & 49d1d64f & C2 & 4be741c5 & S1 & 4c4377d9 & S2 & 4c5c2cf0 & C2 & 50846271 & A2 & 508bd3b6 & S1 & 50cb2852 & S3 \\
5117e062 & C1 & 5168d44c & C1 & 539a4f51 & S3 & 53b68214 & S2 & 543a7ed5 & S3 & 54d82841 & S1 & 54d9e175 & C1 & 5521c0d9 & S1 \\
5582e5ca & amb & 5614dbcf & C1 & 56dc2b01 & S3 & 56ff96f3 & C1 & 57aa92db & S3 & 5ad4f10b & C2 & 5bd6f4ac & S1 & 5c0a986e & C1 \\
5c2c9af4 & S3 & 5daaa586 & A1 & 60b61512 & S3 & 6150a2bd & S1 & 623ea044 & S3 & 62c24649 & S2 & 63613498 & A2 & 6430c8c4 & L1 \\
6455b5f5 & C1 & 662c240a & S2 & 67385a82 & C1 & 673ef223 & S3 & 6773b310 & C2 & 67a3c6ac & S1 & 67a423a3 & S3 & 67e8384a & S2 \\
681b3aeb & C1 & 6855a6e4 & S1 & 68b16354 & S1 & 694f12f3 & S3 & 6a1e5592 & A2 & 6aa20dc0 & S3 & 6b9890af & C2 & 6c434453 & C1 \\
6cdd2623 & S3 & 6cf79266 & C1 & 6d0160f0 & amb & 6d0aefbc & S2 & 6d58a25d & S3 & 6d75e8bb & C1 & 6e02f1e3 & S3 & 6e19193c & C1 \\
6e82a1ae & C1 & 6ecd11f4 & C2 & 6f8cd79b & S3 & 6fa7a44f & S2 & 72322fa7 & A2 & 72ca375d & S3 & 73251a56 & S1 & 7447852a & S1 \\
7468f01a & C2 & 746b3537 & S1 & 74dd1130 & S1 & 75b8110e & C1 & 760b3cac & S3 & 776ffc46 & C1 & 77fdfe62 & C1 & 780d0b14 & S1 \\
7837ac64 & L1 & 794b24be & C1 & 7b6016b9 & S3 & 7b7f7511 & S2 & 7c008303 & C2 & 7ddcd7ec & S3 & 7df24a62 & A2 & 7e0986d6 & S3 \\
7f4411dc & S3 & 7fe24cdd & S2 & 80af3007 & C2 & 810b9b61 & S3 & 82819916 & S1 & 83302e8f & C1 & 834ec97d & S3 & 8403a5d5 & S3 \\
846bdb03 & C2 & 855e0971 & S1 & 85c4e7cd & S3 & 868de0fa & S3 & 8731374e & C2 & 88a10436 & A2 & 88a62173 & C2 & 890034e9 & C1 \\
8a004b2b & C2 & 8be77c9e & S2 & 8d5021e8 & S2 & 8d510a79 & S3 & 8e1813be & S3 & 8e5a5113 & S2 & 8eb1be9a & S2 & 8efcae92 & A2 \\
8f2ea7aa & L1 & 90c28cc7 & S1 & 90f3ed37 & C1 & 913fb3ed & S3 & 91413438 & C1 & 91714a58 & C1 & 9172f3a0 & K1 & 928ad970 & S3 \\
93b581b8 & A2 & 941d9a10 & C1 & 94f9d214 & L1 & 952a094c & S3 & 9565186b & C1 & 95990924 & S3 & 963e52fc & S2 & 97999447 & C1 \\
97a05b5b & A1 & 98cf29f8 & S1 & 995c5fa3 & amb & 99b1bc43 & L1 & 99fa7670 & C1 & 9aec4887 & C2 & 9af7a82c & S3 & 9d9215db & S2 \\
9dfd6313 & S1 & 9ecd008a & S2 & 9edfc990 & C1 & 9f236235 & S1 & a1570a43 & amb & a2fd1cf0 & S3 & a3325580 & S3 & a3df8b1e & S1 \\
a416b8f3 & S2 & a48eeaf7 & S1 & a5313dff & S3 & a5f85a15 & C1 & a61ba2ce & C1 & a61f2674 & A1 & a64e4611 & C1 & a65b410d & S3 \\
a68b268e & C1 & a699fb00 & C1 & a740d043 & S3 & a78176bb & S3 & a79310a0 & C1 & a85d4709 & C1 & a87f7484 & L1 & a8c38be5 & S3 \\
a8d7556c & C1 & a9f96cdd & S3 & aabf363d & C1 & aba27056 & S3 & ac0a08a4 & K1 & ae3edfdc & C1 & ae4f1146 & S3 & aedd82e4 & C1 \\
af902bf9 & S3 & b0c4d837 & C2 & b190f7f5 & S2 & b1948b0a & C1 & b230c067 & C1 & b27ca6d3 & A2 & b2862040 & C1 & b527c5c6 & S3 \\
b548a754 & S3 & b60334d2 & A2 & b6afb2da & S3 & b7249182 & C2 & b775ac94 & A2 & b782dc8a & L1 & b8825c91 & S1 & b8cdaf2b & S3 \\
b91ae062 & K1 & b94a9452 & amb & b9b7f026 & S3 & ba26e723 & C1 & ba97ae07 & S3 & bb43febb & S3 & bbc9ae5d & S3 & bc1d5164 & L1 \\
bd4472b8 & C2 & bda2d7a6 & S3 & bdad9b1f & L1 & be94b721 & C1 & beb8660c & S1 & c0f76784 & S3 & c1d99e64 & C1 & c3e719e8 & amb \\
c3f564a4 & S1 & c444b776 & C2 & c59eb873 & K1 & c8cbb738 & C1 & c8f0f002 & C1 & c909285e & S3 & c9e6f938 & S2 & c9f8e694 & C2 \\
caa06a1f & S2 & cbded52d & C1 & cce03e0d & amb & cdecee7f & amb & ce22a75a & S3 & ce4f8723 & L1 & ce602527 & K1 & ce9e57f2 & S1 \\
cf98881b & C1 & d037b0a7 & S3 & d06dbe63 & A2 & d07ae81c & C1 & d0f5fe59 & S3 & d10ecb37 & amb & d13f3404 & S3 & d22278a0 & S3 \\
d23f8c26 & C1 & d2abd087 & C1 & d364b489 & C1 & d406998b & C1 & d43fd935 & S3 & d4469b4b & amb & d4a91cb9 & S3 & d4f3cd78 & S3 \\
d511f180 & C1 & d5d6de2d & S3 & d631b094 & amb & d687bc17 & C1 & d6ad076f & C2 & d89b689b & C1 & d8c310e9 & S2 & d90796e8 & C1 \\
d9f24cd1 & S3 & d9fac9be & C1 & dae9d2b5 & L1 & db3e9e38 & S1 & db93a21d & L1 & dbc1a6ce & C1 & dc0a314f & S2 & dc1df850 & S3 \\
dc433765 & S1 & ddf7fa4f & A2 & de1cd16c & S3 & ded97339 & S3 & e179c5f4 & S1 & e21d9049 & S2 & e26a3af2 & S1 & e3497940 & S2 \\
e40b9e2f & S2 & e48d4e1a & S1 & e5062a87 & C1 & e509e548 & C1 & e50d258f & S3 & e6721834 & S3 & e73095fd & S3 & e76a88a6 & S3 \\
e8593010 & C1 & e8dc4411 & A1 & e9614598 & S3 & e98196ab & S2 & e9afcf9a & S1 & ea32f347 & C1 & ea786f4a & C1 & eb281b96 & S2 \\
eb5a1d5d & S3 & ec883f72 & S3 & ecdecbb3 & S3 & ed36ccf7 & S1 & ef135b50 & S3 & f15e1fac & S3 & f1cefba8 & C1 & f25fbde4 & S3 \\
f25ffba3 & S2 & f2829549 & L1 & f35d900a & S3 & f5b8619d & S2 & f76d97a5 & C1 & f8a8fe49 & S1 & f8b3ba0a & S1 & f8c80d96 & S3 \\
f8ff0b80 & S3 & f9012d9b & S1 & fafffa47 & L1 & fcb5c309 & S3 & fcc82909 & A2 & feca6190 & S3 & ff28f65a & A2 & ff805c23 & S2 \\
\end{longtable}
\end{scriptsize}

\subsection{Validation Notes}

The classification methodology is described in Section~3.1. Briefly, each task was classified by analyzing its generator code structure, identifying the core transformation primitives, and mapping to the taxonomy. Validation was performed by comparing automated classifications against manual expert review of a stratified sample. The 14 ambiguous tasks exhibited characteristics spanning multiple categories (e.g., tasks requiring both topological reasoning and iterative refinement) and were conservatively excluded from accuracy calculations.

\clearpage
\section{Reproducibility and Data Lineage}
\label{app:reproducibility}

All empirical claims in this paper are programmatically verifiable through our open-source reproduction package. This appendix provides a systematic mapping between key findings and their corresponding source files, enabling complete verification of results.

\subsection{Primary Data Sources}

\begin{description}
  \item[\filepath{paper/champion\_FINAL\_CLEAR.txt}] Complete performance analysis of the champion generalist model on the re-arc validation set, including cell and grid accuracy for all 400 tasks. Referenced throughout Section~7.
  
  \item[\filepath{reproduction/outputs/atomic\_lora\_training\_summary.json}] Training results for all 302 task-specific LoRA fine-tuning experiments, containing epoch-by-epoch metrics. Primary source for Section~7.1 (Compositional Gap analysis).
  
  \item[\filepath{reproduction/outputs/arc\_agi\_2\_experiments/summary/arc\_agi\_2\_cross\_run\_summary.csv}] Aggregated results across 5 random seeds for the champion model evaluated on 120 ARC-AGI-2 public test tasks. Source for Section~7.3 (Generalization experiments).
\end{description}

\subsection{Claim-to-Source Mapping}

The following table maps each major empirical claim to its verification script and source data:

\begin{table}[H]
  \centering
  \scriptsize
  \caption{Reproducibility mapping for key empirical findings (see subsections below for full paths)}
  \label{tab:reproducibility-map}
  \begin{tabular}{@{}p{0.7cm}p{4.8cm}p{4.3cm}p{4.5cm}@{}}
    \toprule
    \textbf{Sec.} & \textbf{Finding / Claim} & \textbf{Verification Script} & \textbf{Source Data} \\
    \midrule
    5.6 & 35.3\% of re-arc tasks are Low/Very Low affinity & fig1\_category\_distribution.py & classification\_summary.txt \\[0.5em]
    \midrule
    7.1.1 & 69.5\% tasks exhibit Compositional Gap ($>$80\% cell, $<$10\% grid) & calculate\_compositional\_gap.py & atomic\_lora\_training\_summary.json \\[0.5em]
    \midrule
    7.1.3 & Champion: 97.6\% cell, 0\% grid on task 31aa019c & See primary data & champion\_FINAL\_CLEAR.txt Sec. 2.7 \\[0.5em]
    \midrule
    7.2.1 & 9/21 A2 tasks: 0\% grid despite 400 examples & verify\_a2\_failures.py & atomic\_lora\_training\_summary.json, tasks\_by\_category.json \\[0.5em]
    \midrule
    7.2.1 & Task 694f12f3: 99.33\% cell, 17.75\% grid plateau & generate\_smoking\_gun\_figure.py & atomic\_lora\_training\_summary.json \\[0.5em]
    \midrule
    7.3.1 & Grid accuracy: 2.34\% $\rightarrow$ 0.34\% (re-arc to AGI-2) & arc\_agi\_2\_analysis.py & champion\_FINAL\_CLEAR.txt, arc\_agi\_2\_cross\_run\_summary.csv \\[0.5em]
    \midrule
    7.3.2 & Cell accuracy: 71.6\% $\rightarrow$ 89.37\% on AGI-2 & generate\_arc\_agi\_2\_comparison.py & Same as above \\[0.5em]
    \midrule
    7.3.3 & 68.6\% AGI-2 failures in Low/Very Low affinity & analyze\_arc\_agi\_2\_failures.py & arc\_agi\_2\_cross\_run\_summary.csv \\[0.5em]
    \midrule
    7.4.1 & S3-A: 5.68\% grid; S3-B: 0.10\% (57$\times$ diff.) & Manual analysis & champion\_FINAL\_CLEAR.txt Sec. 2.5 \\[0.5em]
    \midrule
    7.4.2 & 11/18 S3-A tasks show Comp. Failure pattern & generate\_s3\_performance\_profiles.py & champion\_FINAL\_CLEAR.txt \\[0.5em]
    \bottomrule
  \end{tabular}
\end{table}

\subsection{Configuration and Taxonomy Files}

\begin{description}
  \item[\filepath{atomic\_lora\_training.yaml}] Complete hyperparameter specification for all LoRA fine-tuning experiments (rank 16, alpha 32, 200 epochs with early stopping).
  
  \item[\filepath{split\_manifest.json}] Defines the 308 training / 92 validation split used throughout this work.
  
  \item[\filepath{reproduction/data/taxonomy/tasks\_by\_category.json}] Complete mapping of all 400 re-arc tasks to their taxonomy categories.
  
  \item[\filepath{reproduction/data/taxonomy/s3\_final\_classification.json}] Sub-classification of 108 S3 tasks into S3-A (pattern-based, n=77) and S3-B (graph reasoning, n=31).
  
  \item[\filepath{/paper/arc\_task\_taxonomy.md}] Canonical definitions and theoretical basis for all 9 taxonomy categories.
\end{description}

\subsection{Figure Generation Scripts}

All figures in this paper are programmatically generated from source data. The following scripts produce the main figures:

\begin{description}
  \item[Figure~1] \filepath{figures/fig1\_category\_distribution.py} — Category distribution with affinity color-coding
  \item[Figure~\ref{fig:gap-sensitivity}] \filepath{paper/generate\_compositional\_gap\_sensitivity.py} — Sensitivity heatmap for compositional gap thresholds
  \item[Figure~\ref{fig:arc-agi2-comparison}] \filepath{scripts/generate\_arc\_agi\_2\_comparison.py} — Two-panel comparison of re-arc vs ARC-AGI-2 performance
  \item[Figure~\ref{fig:failures-concentrate}] \filepath{scripts/generate\_failure\_concentration.py} — Failure concentration by affinity level on ARC-AGI-2
  \item[Figure~\ref{fig:s3a-profiles}] \filepath{scripts/generate\_s3\_performance\_profiles.py} — S3-A performance scatter plot
\end{description}

\subsection{Additional Documentation}

\begin{description}
  \item[\filepath{reproduction/docs/methodology/s3\_generator\_analysis.md}] Rationale and code excerpts justifying the S3-A/S3-B sub-classification based on generator function analysis.
  
  \item[\filepath{reproduction/docs/results/arc\_agi\_2\_evaluation.md}] Complete documentation of ARC-AGI-2 evaluation protocol and results across 5 seeds.
  
  \item[\filepath{reproduction/docs/methodology/ambiguous\_tasks\_analysis.md}] Analysis of 14 ambiguous tasks that could not be automatically classified by the rule-based system.
\end{description}

\textbf{Repository Structure:} All file paths referenced in this appendix are relative to the root of our public reproduction package. Complete setup instructions and environment specifications are provided in the repository README.
    
\end{document}